\documentclass[lettersize,journal]{IEEEtran}
\usepackage{amsmath,amsfonts}
\usepackage{algorithmic}
\usepackage{algorithm}
\usepackage{array}
\usepackage[caption=false,font=normalsize,labelfont=sf,textfont=sf]{subfig}
\usepackage{textcomp}
\usepackage{stfloats}
\usepackage{url}
\usepackage{verbatim}
\usepackage{graphicx}
\usepackage{cite}
\usepackage[table,xcdraw]{xcolor}
\usepackage{booktabs}
\usepackage{multirow}
\usepackage{makecell}
\usepackage{longtable}
\usepackage{supertabular}
\usepackage{hyperref}
\DeclareSubrefFormat{myparens}{#1(#2)}
\usepackage{tikz,xcolor,hyperref}% Make Orcid icon
\definecolor{lime}{HTML}{A6CE39}
\DeclareRobustCommand{\orcidicon}{%
    \begin{tikzpicture}
    \draw[lime, fill=lime] (0,0) 
    circle [radius=0.16] 
    node[white] {{\fontfamily{qag}\selectfont \tiny ID}};    \draw[white, fill=white] (-0.0625,0.095) 
    circle [radius=0.007];    \end{tikzpicture}
    \hspace{-2mm}}
\foreach \x in {A, ..., Z}{%
    \expandafter\xdef\csname orcid\x\endcsname{\noexpand\href{https://orcid.org/\csname orcidauthor\x\endcsname}{\noexpand\orcidicon}}
}
\newcommand{\etal}{\textit{et al}.}

\bstctlcite{IEEEexample:BSTcontrol}

\begin{document}

\title{Exploring Radar Data Representations in Autonomous Driving: A Comprehensive Review}

\author{
%Shanliang Yao\textsuperscript{\orcidA{}},
%Runwei Guan\textsuperscript{\orcidB{}}, 
%Yong Yue\textsuperscript{\orcidI{}}, \\ 

%Eng Gee Lim\textsuperscript{\orcidH{}}, \IEEEmembership{Senior Member,~IEEE}, 
%Hyungjoon Seo\textsuperscript{\orcidK{}}, 
%Ka Lok Man\textsuperscript{\orcidJ{}}, 
%Xiaohui Zhu\textsuperscript{\orcidL{}}, 
%Yutao Yue\textsuperscript{\orcidM{}}
Shanliang Yao,
Runwei Guan, 
Zitian Peng,
Chenhang Xu,
Yilu Shi, 
Weiping Ding, \IEEEmembership{Senior Member,~IEEE},\\
Eng Gee Lim, \IEEEmembership{Senior Member,~IEEE}, 
%Weiping Ding\textsuperscript{3}, \IEEEmembership{Senior Member,~IEEE}, 
Yong Yue, 
Hyungjoon Seo, 
Ka Lok Man,
Jieming Ma, \\ 
Xiaohui Zhu, 
Yutao Yue

\thanks{$^{\text{1}}$ Shanliang Yao is with the School of Information Engineering, Yancheng Institute Technology, Yancheng 224051, China. (email: shanliang.yao@ycit.edu.cn).}
\thanks{$^{\text{2}}$ Runwei Guan and Yutao Yue are with Thrust of Artificial Intelligence and Thrust of Intelligent Transportation, The Hong Kong University of Science and Technology (Guangzhou), Guangzhou 511400, China. (email: runwei.guan@liverpool.ac.uk, yutaoyue@hkust-gz.edu.cn).}
\thanks{$^{\text{3}}$ Zitian Peng and Hyungjoon Seo are with Faculty of Science and Engineering, University of Liverpool, Liverpool, UK. (email: \{pszpeng, hyungjoon.seo\}@liverpool.ac.uk).}
\thanks{$^{\text{4}}$ Chenhang Xu, Yilu Shi, Yong Yue, Eng Gee Lim, Ka Lok Man, Jieming Ma and Xiaohui Zhu are with School of Advanced Technology, Xi'an Jiaotong-Liverpool University, Suzhou, China. (email: \{chenhang.xu21, yilu.shi20\}@student.xjtlu.edu.cn. \{yong.yue, enggee.lim, ka.man, jieming.ma, xiaohui.zhu\}@xjtlu.edu.cn).}
\thanks{$^{\text{5}}$ Weiping Ding is with School of Information Science and Technology, Nantong University, Nantong 226019, China. (email: dwp9988@163.com).}
\thanks{$^{\dagger}$ Corresponding author: xiaohui.zhu@xjtlu.edu.cn, yutaoyue@hkust-gz.edu.cn}
}

\maketitle

\begin{abstract}
With the rapid advancements of sensor technology and deep learning, autonomous driving systems are providing safe and efficient access to intelligent vehicles as well as intelligent transportation.
Among these equipped sensors, the radar sensor plays a crucial role in providing robust perception information in diverse environmental conditions. 
This review focuses on exploring different radar data representations utilized in autonomous driving systems. Firstly, we introduce the capabilities and limitations of the radar sensor by examining the working principles of radar perception and signal processing of radar measurements.
Then, we delve into the generation process of five radar representations, including the ADC signal, radar tensor, point cloud, grid map, and micro-Doppler signature.
For each radar representation, we examine the related datasets, methods, advantages and limitations. Furthermore, we discuss the challenges faced in these data representations and propose potential research directions.
Above all, this comprehensive review offers an in-depth insight into how these representations enhance autonomous system capabilities, providing guidance for radar perception researchers.
To facilitate retrieval and comparison of different data representations, datasets and methods, we provide an interactive website at \url{https://radar-camera-fusion.github.io/radar}.

\end{abstract}

\begin{IEEEkeywords}
Radar perception, autonomous driving, data representations, intelligent vehicles, intelligent transportation.
\end{IEEEkeywords}

\section{Introduction}

% 雷达在自动驾驶中的作用、优势
Nowadays, the automotive industry has witnessed significant advancements in autonomous driving technologies, revolutionizing intelligent vehicles and intelligent transportation. 
With the aim of creating safer and more efficient roadways, one crucial aspect is the development of reliable sensor perception for autonomous driving systems. 
Compared to LiDARs and cameras, the fundamental advantage of radar sensors lies in their capabilities in range, velocity and angle measurements \cite{yao2023radar, venon2022millimeter}. 
Additionally, radars have robust sensing capabilities and exhibit superior effectiveness under adverse lighting and weather conditions \cite{major2019vehicle, wang2021rodnet}. 
Moreover, radar sensors can even perceive objects behind walls, providing vehicles with early warning of potential obstacles or hazards \cite{scheiner2020seeing, li2022detection}. 
All of the above characteristics make radar an indispensable component in ensuring the reliability and superiority of autonomous driving systems, especially in scenarios where other sensor inputs might be limited or compromised.

\begin{figure}[!t]
\begin{center}
   \includegraphics[width=1\linewidth]{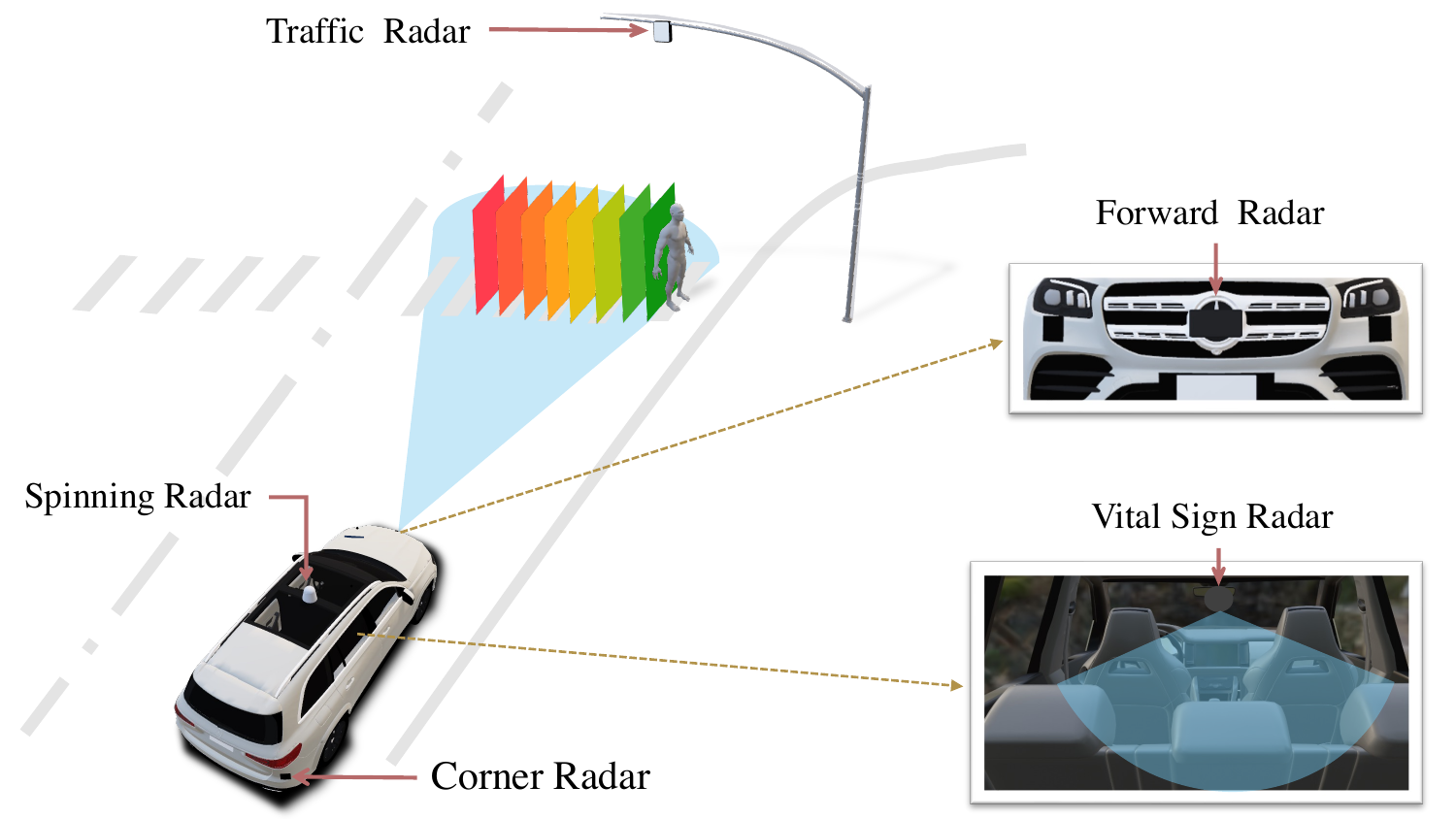}
\end{center}
   \caption{Radar perception in autonomous driving. (a) Radars on the vehicle are employed to detect objects in the path and surroundings. (b) Radars inside the cabin are leveraged to monitor occupant vital signs and behaviors. (c) Radars alongside roadways are utilized to measure the speed of passing vehicles and estimate traffic flow.}
\label{fig:cover}
\end{figure}

% 引入雷达传感器，细化的应用（1、对抗天气；2、对抗光线；3、远距离和墙后目标；缺点：1、稀疏；2、无语义信息）
% 雷达挑战（标定、点不准、杂波噪音、标注）

% 雷达的应用
As depicted in Fig. \ref{fig:cover}, radar sensors have extensive applications in autonomous driving, serving various purposes in the perception of the surrounding environment.
Radars mounted on the roof, front and angular directions of an intelligent vehicle are utilized to detect objects in the path and surroundings, providing essential information for path planning and adaptive cruise control \cite{stroescu2019classification, stroescu2021object, karangwa2023vehicle}. 
The radar inside the cabin is leveraged to monitor occupant vital signs, left-behind children and driver behaviors (e.g., drowsiness, dangerous maneuvers, health emergencies) in a privacy-protected manner, allowing the autonomous driving system to make alerts as well as safety measures \cite{zhang2023overview, da2019theoretical, wang2021driver, chen2022attention}. 
Furthermore, radar systems deployed alongside roadways can measure the speed of passing vehicles and estimate traffic flow. These traffic data are valuable for traffic optimization, congestion management and intelligent transportation systems, leading to improved road safety and smoother traffic operations \cite{jiang2022improve, huang2020radar, emami2021long}.

%
%This review aims to explore different radar representations employed in autonomous driving and their impact on perception tasks. By examining the various representation techniques, advantages, and limitations, 
%we seek to provide an in-depth analysis of how these representations contribute to enhancing the capabilities of autonomous vehicles. 
%By studying radar perception in the context of autonomous driving, we aim to gain insight into the state-of-the-art approaches and emerging trends in this field. 
% 本文的内容（原理、数据集、方法、讨论）

\begin{figure*}[htbp]
\begin{center}
   \includegraphics[width=0.95\linewidth]{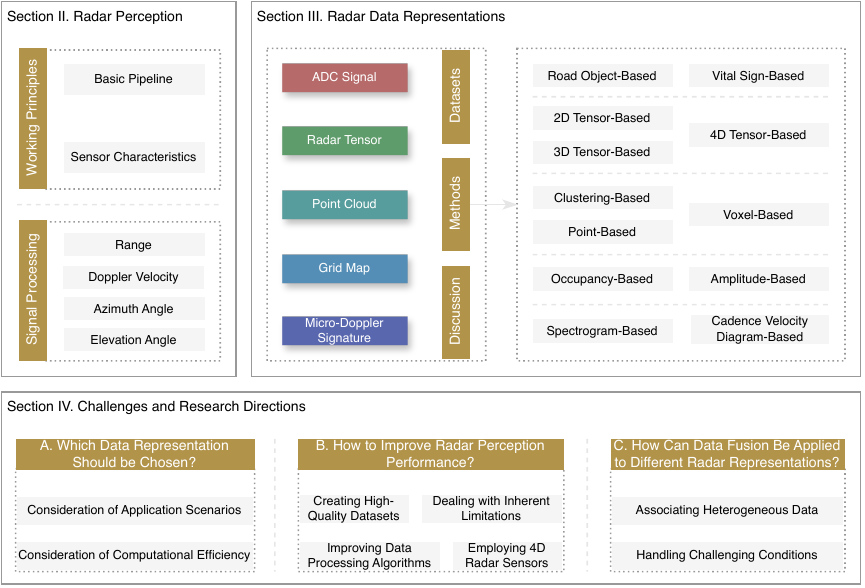}
\end{center}
   \caption{Overview of this review. Section \ref{sec:Radar Perception} provides an overview of radar perception, including its working principles and signal processing techniques. 
Section \ref{sec:Radar Data Representations} presents an in-depth examination on datasets and methods of different radar data representations. 
Section \ref{sec:Discussion} discusses the challenges and potential directions for research and development for radar perception in autonomous driving.
}
\label{fig:overview}
\end{figure*}

% 雷达的表征
% 表征多，对研究者来说难以区分、不知道各自的优缺点，什么时候该用啥，不知道算法的发展进展。
As far as radar data is concerned, various application scenarios and radar types provide different data representations. Specifically, each representation has distinct characteristics and challenges, involving different data processing methods and network architectures. 
Moreover, each representation aims to extract needed information from raw radar measurements and convert it into a suitable format for downstream perception tasks. 
With limited focus on radar data representations in existing reviews, it is challenging for researchers to comprehensively understand this emerging autonomous driving field.
This review attempts to narrow this gap by exploring five radar representations (i.e., ADC signal, radar tensor, point cloud, grid map, micro-Doppler signature) leveraged in autonomous driving and their impact on perception tasks. Specifically, we provide guidance for researchers to differentiate between each representation, understand the advantages and limitations of each, and select the appropriate representation and algorithm for the specific task.
Our review offers the following key contributions:

\begin{itemize}
\item To the best of our knowledge, this is the first review that explores different data representations for radar perception in autonomous driving.
\item We offer an up-to-date (2019-2024) overview of radar-based datasets and algorithms, providing in-depth research on their advantages and limitations.
\item We analyze the significant challenges and open questions related to these data representations, and propose potential research directions for further investigation.
\item We develop an interactive and regularly updated website that facilitates easier retrieval and comparison of the datasets and methods.
%\item facilitating easy retrieval and comparison of datasets and methodologies for researchers.
\end{itemize}

The remainder of this review is structured in Fig. \ref{fig:overview} and is described as follows: 
Section \ref{sec:Radar Perception} provides an overview of radar working principles and signal processing techniques, serving as the foundation for generating various radar data representations.
Section \ref{sec:Radar Data Representations} presents an in-depth examination on datasets and methods for each radar data representation. Additionally, this section discusses the advantages and limitations of each radar data representation, assisting readers in selecting the most suitable approach for their research.
Section \ref{sec:Discussion} delves into challenges and potential directions in research and development for radar perception in autonomous driving, guiding readers to focus on current hot topics and explore feasible solutions.
Lastly, Section \ref{sec:Conclusion} summarizes our study and presents an outlook for future works, inspiring researchers to make further strides in radar perception.

\section{Radar Perception}\label{sec:Radar Perception}

%The paper examines the fundamental principles of radar technology and provides an overview of its key components, including antennas, transceivers, and digital signal processing units. Furthermore, it delves into the challenges of radar perception in the autonomous driving domain, such as clutter, multi-path reflections, and object identification and tracking. Various signal processing algorithms and methodologies to mitigate these challenges are discussed, including noise filtering, target detection, and radar data fusion.

In this section, we provide an overview of the underlying principles of radar technology and how it operates within the autonomous driving system. We delve into the physics behind radar sensing, including the transmission and reception of radio waves, as well as the mechanisms for extracting valuable information from the returned signals. Understanding these principles is crucial for comprehending the capabilities of radar perception and the generation process of various radar data representations.

\subsection{Working Principles}

% 雷达背景
Radar sensors play crucial roles in enabling autonomous systems to sense vehicle occupants, vehicle surroundings, and the traffic environment. The term ``radar" stands for ``Radio Detection and Ranging", which emits millimeter waves that bounce off objects and return to the sensor. This operation provides information about the object's location, relative velocity, and internal characteristics \cite{iovescu2017fundamentals}.
In the following, we introduce the basic radar working pipeline and sensor characteristics derived from these principles.

\begin{figure}[htbp]
\begin{center}
   \includegraphics[width=1\linewidth]{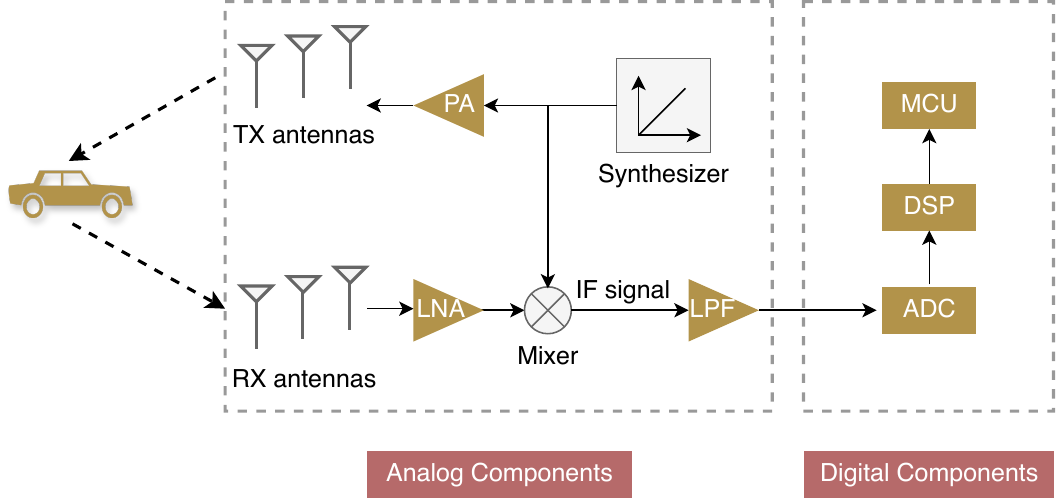}
\end{center}
   \caption{Overview of radar working pipeline.}
\label{fig:radar}
\end{figure}

% 雷达构成，工作过程
\paragraph{Basic Pipeline}
Fig. \ref{fig:radar} demonstrates a basic radar system that comprises various analog components, including a synthesizer, a Power Amplifier (PA), transmitters (TX antennas), receivers (RX antennas), a Low Noise Amplifier (LNA), a mixer and a Low Pass Filter (LPF), as well as digital components such as an Analog-to-Digital Converter (ADC), a Digital Signal Processor (DSP) and a MicroController Unit (MCU). 
When the radar starts working, first, the synthesizer generates a linear frequency-modulated pulse called a ``chirp", which is amplified in power and transmitted by the TX antenna. Second, the RX antenna captures the reflected chirp of the target that is amplified with low noise. Third, by combining the RX and TX signals, the mixer produces an Intermediate Frequency (IF) signal, which is then converted into digital values via the ADC. 
Generally, a radar system contains multiple TX and RX antennas, resulting in multiple IF signals. Information about the target object, such as range, Doppler velocity and azimuth angle, is contained in these IF signals, which can be separated by the DSP using embedding different signal processing algorithms \cite{saponara2019radar}. Based on extracted target information from the DSP and dynamic information from the radar, the MCU serves as a computer to evaluate system requirements and make informed decisions.

%天线是用来发射和接收毫米波信号的，一个毫米波雷达由多个发射天线和多个接收天线组成。
%前端收发组件是毫米波雷达的核心射频部分，负责毫米波信号调制、发射、接收以及回波信号的解调.
%数字信号处理系统通过嵌入不同的信号处理算法，提取从前端采集得到的中频信号，获得特定类型的目标信息。
%MCU根据DSP输出的目标信息，结合车身动态信息进行数据融合，最终对车辆前方出现的障碍物进行分析判断，并迅速做出处理和发出指令。

% 工作原理：测距：TOF，测速：多普勒，测方位：天线阵列

%Based on the Time of Flight (TOF) principle, the radar sensor calculates the range from the object by the time difference between the transmitted and reflected signals. Based on the Doppler principle, when there is a relative movement between the emitted electromagnetic wave and the detected target, the frequency of the returned wave differs from that of the emitted wave. Thus, the target's relative Doppler velocity to the radar can be measured using this frequency difference. 
%Leveraging the array signal processing method, the azimuth angle is calculated by the phase difference between the chirps reflected from the parallel receivers. Since the receivers of traditional 3D (range, doppler, azimuth angle) radar sensors are only lined up in a 2D direction, targets are only detected in 2D horizontal coordinates without vertical height information. Recently, with advancements in radar technologies, 4D (range, doppler, azimuth angle, elevation angle) radar sensors have been developed with antennas arranged horizontally and vertically, enabling the measurement of elevation angle.

Utilizing the Time of Flight (TOF) principle, radar sensors determine the range of an object by calculating the time difference between the transmitted and reflected signals. Additionally, based on the Doppler principle, radar sensors can measure the relative Doppler velocity of a target by analyzing the frequency difference between the emitted and received waves, capturing any relative movement between the radar and the target.
To determine the azimuth angle, an array signal processing method is employed, which involves calculating the phase difference between the chirps reflected from parallel receivers. However, traditional 3D radar sensors with receivers arranged in a 2D manner only provide detection in horizontal coordinates, lacking vertical height information.
Recent advancements in radar technologies have led to the development of 4D radar sensors, incorporating antennas arranged both horizontally and vertically \cite{li2019pioneer, han20234d}. This configuration allows for the measurement of the elevation angle, thereby enabling the capture of elevation information in addition to the range, Doppler velocity, and azimuth angle measurements.

In pipeline systems, the number of steps is a critical factor that directly impacts complexity, performance, modularity, and maintainability. A larger number of steps enhances modularity by breaking down complex tasks into smaller, more manageable components, which improves maintainability and facilitates debugging. However, an excessive number of steps can increase computational overhead, introduce latency, and complicate system integration. These trade-offs must be carefully balanced based on the specific requirements of the application.

% 从原理的角度引入雷达优点：对抗恶劣天气、全天候工作。

% 从原理的角度引入雷达缺点：角度、高度精度低（点不准），点稀疏，静止障碍物感知能力弱
\paragraph{Sensor Characteristics}
With the basic capabilities of measuring range, Doppler velocity, azimuth angle and elevation angle, radar can determine the location of obstacles, allowing vehicles to make informed decisions and navigate safely.
Unlike other light-wave-based sensors, radar sensors emit radio waves at longer wavelengths. This characteristic allows radar waves to penetrate fog, rain, snow, smoke, and dust \cite{appleby2007millimeter}. Consequently, radar systems can reliably detect and measure distances to objects even in severe weather conditions, making them highly dependable in a wide range of real-world scenarios.
Moreover, radar waves can penetrate certain materials (e.g., walls, vegetation) and reflect off the hidden objects, enabling the sensor to detect objects situated around corners or obstructed by other obstacles \cite{scheiner2020seeing}. This unique characteristic enhances the detection capabilities of the system, particularly in complex urban environments where numerous obstacles may impede direct line of sight.

While radar sensors offer numerous advantages, it is essential to consider their inherent limitations.
They exhibit limited angular resolution, making it challenging to differentiate between closely located objects. Additionally, sparse point clouds generated by radars, with only a few points on pedestrians and a small number on cars, are insufficient for accurately outlining object contours and extracting detailed geometric information \cite{fritsche2016radar, liu2022deep}.
While radar measurements provide the radial velocity component, they lack information about tangential velocity. This limitation makes it difficult to accurately estimate the complete motion of objects in dynamic scenes \cite{long2021full, li2022mathsf}.
More importantly, radar data is susceptible to noise from various sources, including multi-path interference, electrical interference, and equipment imperfections \cite{griffiths2014radar, kopp2023tackling, zhou2022towards}. This noise impacts the precision and reliability of radar measurements, potentially leading to false detections.
%Furthermore, radars are weak in the perception of stationary obstacles. Moving targets can be distinguished from the surrounding scene in one dimension of range and velocity. However, radars are highly sensitive to metal, often resulting in strong reflections from stationary objects such as manhole covers on the ground. As a result, stationary objects are usually filtered in practice, resulting in a lack of detecting stationary obstacles.

\subsection{Signal Processing} \label{Signal Processing}

In this section, we review the radar signal processing for radar parameters, including range, Doppler velocity, azimuth angle and elevation angle. 
Subsequently, Radar Cross-Section (RCS) measurement is analyzed to approximate the target's size, shape and material composition, thereby characterizing its reflection properties.
To filter out clutter during the radar signal processing stage, we examine the workflow of Constant False Alarm Rate (CFAR) processing and explore representative CFAR processors. 

%\subsection{Signal Processing}

\subsubsection{Parameter Estimation}

\begin{figure}[h]
\begin{center}
   \includegraphics[width=1\linewidth]{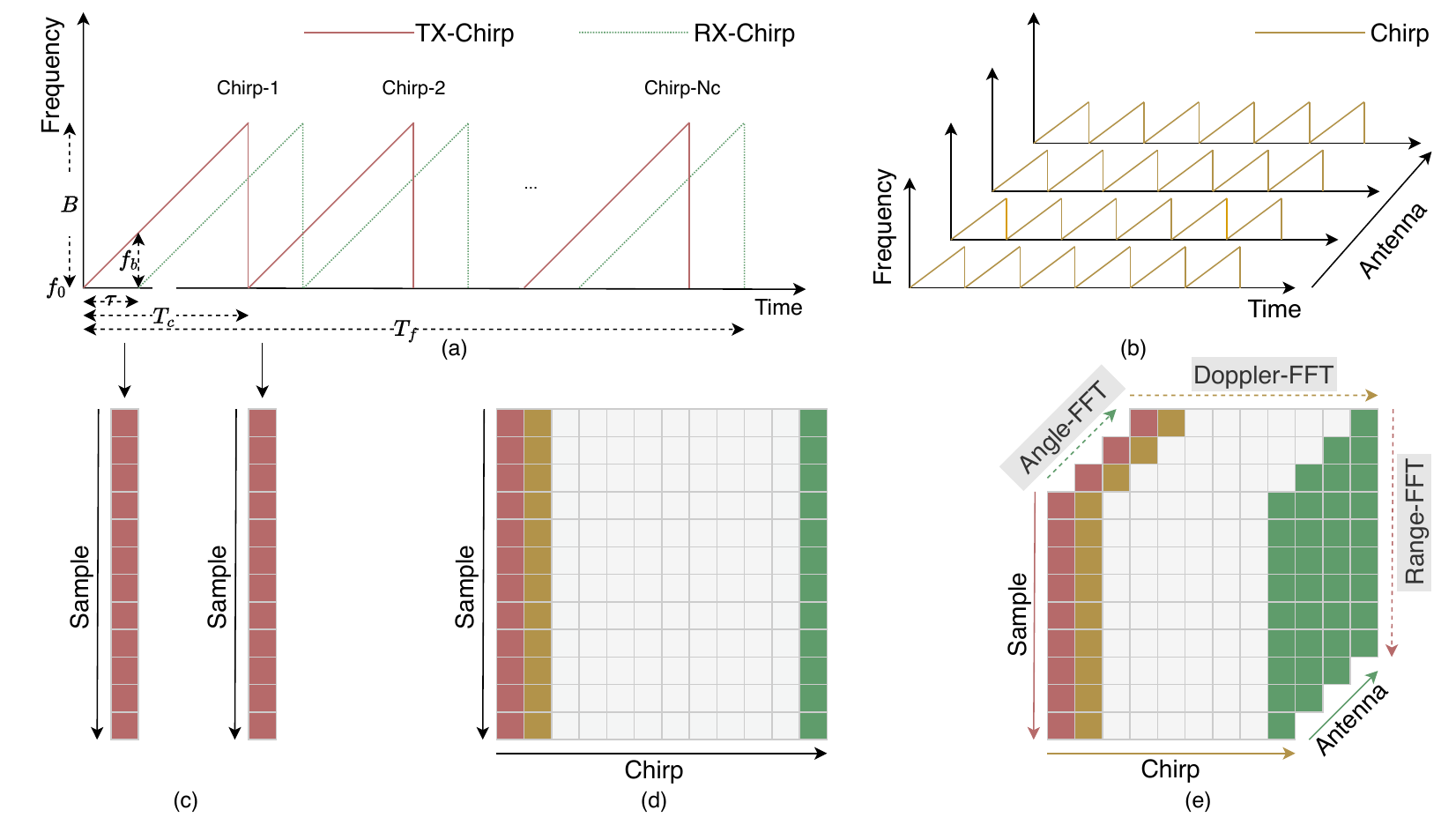}
\end{center}
   \caption{Pipeline of radar signal processing. (a) Frequency of chirps emitted by a TX antenna and received by an RX antenna. (b) Frequency of chirps received by multiple RX antennas. (c) Sampling performed on each chirp. (d) Sample-Chirp map generated from sampling on all chirps. (e) Simple-Chirp-Antenna tensor generated from Sample-Chirp maps based on multiple RX antennas.}
\label{fig:ADC}
\end{figure}

\begin{table*}[htp]
\caption{Summary of radar parameter estimation \cite{iovescu2017fundamentals}} 
\label{tab:parameter}
\footnotesize
\centering
\begin{tabular}{p{2cm}p{3.8cm}p{4.2cm}p{6.4cm}}
\toprule

\multicolumn{1}{l}{\textbf{Parameter}} & \multicolumn{1}{c}{\textbf{Estimation}} & \multicolumn{1}{c}{\textbf{Resolution}} & \multicolumn{1}{c}{\textbf{Notes}}\\ 
\\\midrule

{\vspace{0.4cm}}
Range 
& 
{\vspace{0cm}}
\begin{equation}
{R} =\frac{c f_{b}}{2S} 
\label{equ:range-estimation}
\end{equation} 
& 
{\vspace{0.1cm}}
\begin{equation}
  {R_{res}}= \frac{c}{2B} 
 \label{equ:range-resolution}
\end{equation}
& 
\begin{minipage}[t]{0.35\textwidth}
\begin{itemize}
\setlength{\parsep}{0pt}
\setlength{\parskip}{0pt}
\setlength{\leftmargin}{0pt}
  \item ${c}$ is the speed of the light ($3*10^{8} m/s$).
  \item ${f_{b}}$ is the instantaneous frequency difference at the mixers from TX-chirp and RX-chirp.
  \item ${S}$ is the slope of a chirp.
  \item ${B}$ refers to the bandwidth.
\end{itemize}
\end{minipage}\vspace{0.03cm}
\\\midrule
{\vspace{0.2cm}}
Doppler Velocity
&
\begin{equation}
{V}=\frac{\lambda \Delta \phi}{4\pi T_{c}}
\label{equ:velocity-estimation}
\end{equation}
&
\begin{equation}
%{V_{res}}=\frac{\lambda}{2N_{c}T_{c}} 
{V_{res}}=\frac{\lambda}{2T_{f}}
\label{equ:velocity-resolution}
\end{equation}
& 
%{\vspace{0.005cm}}
\begin{minipage}[t]{0.4\textwidth}
\begin{itemize}
\setlength{\parsep}{0pt}
\setlength{\parskip}{0pt}
\setlength{\leftmargin}{0pt}
  \item $\lambda$ is the wavelength of the transmitted signal.
  \item ${\Delta\phi}$ is a shift in phase.
  \item ${T_{c}}$ is the duration between the chirps.
  \item ${T_{f}}$ is cycle time of a frame.
%  \item ${N_{c}}$ is the number of chirps in one frame
\end{itemize}
\end{minipage}
\\\midrule
{\vspace{0.2cm}}
Azimuth Angle &
\begin{equation}
\theta = sin^{-1}(\frac{\lambda \Delta \phi}{2\pi l_a})
\label{equ:azimuth-estimation}
\end{equation}
&
\begin{equation}
%\theta = sin^{-1}(\frac{\lambda \Delta \phi}{2\pi l})
\theta_{res}=\frac{\lambda}{N_{a}l_{a}cos(\theta)}
\label{equ:azimuth-resolution}
\end{equation}
&
{\vspace{0.05cm}}
\begin{minipage}[t]{0.35\textwidth} \begin{itemize}
\setlength{\parsep}{0pt}
\setlength{\parskip}{0pt}
\setlength{\leftmargin}{0pt}
\item $l_a$ is the length between RX antennas in azimuth.
\item $N_{a}$ is the number of RX antennas in azimuth.
\end{itemize} \end{minipage}
\\\midrule
{\vspace{0.2cm}}
Elevation Angle &
\begin{equation}
\theta = sin^{-1}(\frac{\lambda \Delta \phi}{2\pi l_{e}})
\label{equ:elevation-estimation}
\end{equation}
&
\begin{equation}
%\theta = sin^{-1}(\frac{\lambda \Delta \phi}{2\pi l})
\theta_{res}=\frac{\lambda}{N_{e}l_{e}cos(\theta)}
\label{equ:elevation-resolution}
\end{equation}
&
{\vspace{0.05cm}}
\begin{minipage}[t]{0.4\textwidth} \begin{itemize}
\setlength{\parsep}{0pt}
\setlength{\parskip}{0pt}
\setlength{\leftmargin}{0pt}
\item $l_e$ is the length between RX antennas in elevation.
\item $N_{e}$ is the number of RX antennas in elevation.
\end{itemize} \end{minipage}
\\
\bottomrule
\end{tabular}
\end{table*}

% 距离估计：时间差*速度
% 速度估计：多普勒效应
% 方位角估计：到达不同接收天线的角度

Frequency-Modulated Continuous Wave (FMCW) is a special millimeter wave technology employed in autonomous driving that continuously transmits frequency-modulated signals to measure the attributes of objects. 
As is described in Fig. \ref{fig:ADC}(a), the frequency of the chirp emitted by the TX antenna increases linearly over time, leading to the corresponding frequency of the chirp reflected by the RX antenna.
The IF signal containing information on objects is determined by calculating the difference in the instantaneous frequency of the TX-chirp and RX-chirp, expressed by the variable ${f_{b}}$. 
In the following, we introduce the estimation of each parameter and summarize the mathematical formulation in Table \ref{tab:parameter}.

\paragraph{Range}
As depicted in Fig. \ref{fig:ADC}(c), sampling is initially conducted among each chirp signal. Then, the Fast Fourier Transform (FFT) is performed to produce a spectrum with different peaks, representing different objects at different ranges. Given that FFT operation is utilized to obtain range values, it is also called ``Range-FFT". From Equation \ref{equ:range-estimation}, we know that the range of a target is determined by the difference in the frequency and the slope of a chirp.
Range resolution refers to the capability of distinguishing and resolving two objects that are located very close together along the range dimension. According to Fourier transform theory, the range resolution can be improved by extending the IF signal \cite{cuomo1999ultrawide}. Extending the IF signal requires increasing the signal bandwidth, which results in range resolution being proportional to the signal bandwidth ($B$), as shown in Equation \ref{equ:range-resolution}.

\paragraph{Doppler Velocity}
To measure the Doppler velocity of an object using radar, a common technique involves transmitting two chirps that are separated by a time interval denoted as ${T_{c}}$. For each chirp, the spectrum after Range-FFT peaks at the same position but with different phases. The difference of the measured phase ${\Delta\phi}$ contains the velocity of an object, as noted in Equation \ref{equ:velocity-estimation}. 
For multiple objects at the same range, it is difficult to identify their velocities as the peaks are located at the same position. Thus, a sequence of signals consisting of $N_{c}$ chirps is employed, formulating a Simple-Chirp map described in Fig. \ref{fig:ADC}(d). Doppler-FFT is then performed on the phasors generated by Range-FFT to separate objects since the phase differences between consecutive chirps are different. 
Velocity resolution is the minimum difference in velocity at which a radar can distinguish between two targets at the same range. As expressed in Equation \ref{equ:velocity-resolution}, velocity resolution is inversely proportional to the cycle time of a frame denoted as $T_{f}$.

\paragraph{Azimuth Angle}
Estimating the azimuth angle is accomplished using at least two RX antennas separated by $l_a$. The resulting difference in distances from the target to each RX antenna causes a phase change in the FFT spectrum, from which the Direction of Arrival (DoA) can be obtained, as outlined in Equation \ref{equ:azimuth-estimation}.
To measure the azimuth angle of multiple objects located at the same range and moving at the same velocity, multiple RX antennas are needed, as shown in Fig. \ref{fig:ADC}(b). In this way, the radar signal is described in a  Simple-Chirp-Antenna dimension, as illustrated in Fig. \ref{fig:ADC}(e). Then, an Angle-FFT is performed on the phasor sequences corresponding to the peaks after 2D-FFT (Range-FFT and Doppler-FFT) to resolve azimuth angles.
Azimuth angle resolution $\theta_{res}$ is the minimum angle separation for two objects appearing as separated peaks in the spectrum after Angle-FFT. It can be justified from Equation \ref{equ:azimuth-resolution} that the azimuth angle resolution is maximum when measured perpendicular to the radar system's axis ($\theta=0$). Additionally, it is necessary to increase the number of RX antennas to enhance azimuth angle resolution.

\paragraph{Elevation Angle}
%hardware increasing the number of TX-RX pairs or the aperture of antennas 
%software virtual aperture imaging Super-resolution \cite{han20234d}
Similar to azimuth angle estimation, the measurement of elevation angle requires a minimum of two RX antennas, each separated by a specific length, denoted as $l_e$. 
The mathematical formulation that describes this azimuth angle estimation is expressed in Equation \ref{equ:elevation-estimation}. 
Additionally, the resolution of elevation angle can be calculated using Equation \ref{equ:elevation-resolution},  which can be improved by increasing the number of RX antennas.

\subsubsection{RCS Measurement}

% RCS定义：反映物体反射雷达信号的能力

RCS denotes the ability of an object to reflect a radar signal, and a higher RCS value corresponds to an increased likelihood of detection \cite{richards2010principles}. The value of RCS is expressed as an area in $m^{2}$ \cite{knott2004radar}. However, this value does not simply represent the surface area of the object being detected, but depends on multiple factors, including the target's material, physical geometry, and exterior features of the target, as well as the direction and frequency of the illuminating radar.

% RCS计算：散射功率密度/入射功率密度
In terms of mathematical calculations, RCS is a metric that quantifies the ratio between the scattered density in the direction of the radar and the power density intercepted by the object. Since the power is distributed over a sphere, only a small part of this ($4\pi r^2$) can be received by the radar. Hence, the expression for RCS takes the form in Equation \ref{equ:rcs}:
\begin{equation}
  {\sigma}=\frac{4\pi r^2 S_r}{S_t},
  \label{equ:rcs}
\end{equation}
where $r$ refers to the range between the radar and the target, $S_r$ is the scattered power density at the radar, and $S_t$ represents the incident power density the target intercepts.

\subsubsection{CFAR Processing}

% 杂波定义：内部接收器噪声、外部杂波、干扰的不需要的信号
The radar sensors not only receive the reflected signals from the objects of interest, but also encounter the internal receiver noise and external interfering signals. Signals generated by these unwanted sources are commonly called clutter. Traditional methods like removing signals with zero Doppler and fixing signal thresholds have drawbacks that lead to false alarms.
Consequently, dynamic thresholding, which involves using adaptive threshold values, is crucial in mitigating false alarms and spurious radar detections attributable to noise signals. Compared to fixed thresholds, varied thresholds are involved in reducing false alarms and erroneous radar detections caused by noise. CFAR is the most commonly used method of dynamic thresholding \cite{rohling1983radar, gandhi1988analysis}, facilitating the radar system to autonomously adapt its sensitivity threshold in response to variations in the amplitude of external interference, thus ensuring a consistent level of false alarm probability.

\begin{figure}[h]
\begin{center}
   \includegraphics[width=1\linewidth]{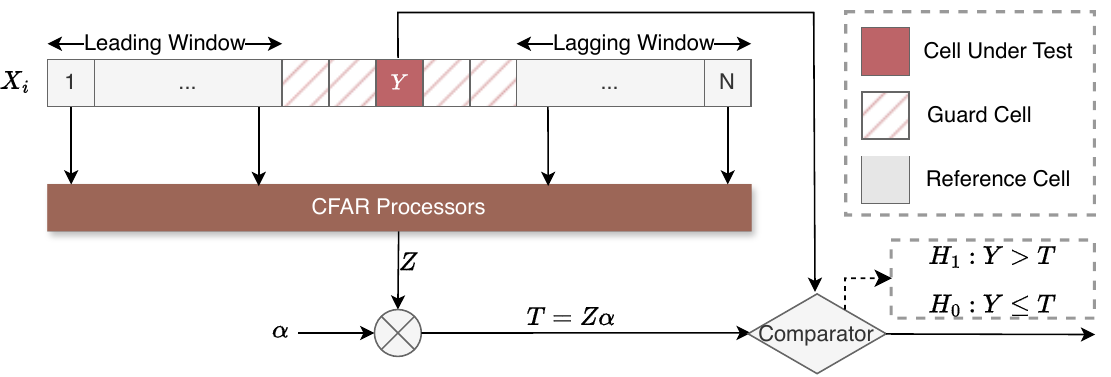}
\end{center}
   \caption{Overview of CFAR processing. Set of $X_1, X_2, ..., X_N$ represents the detection window, $Y$ is the value of CUT, $Z$ represents the clutter background level of the CUT, $T$ is the detection threshold, $\alpha$ donates a scaling factor, $H_1$ declares that an object is located within the CUT, $H_0$ indicates that there is no object in the current CUT.}
   \label{fig:CFAR}
\end{figure}

Fig. \ref{fig:CFAR} describes the overview processing flow of CFAR processors. 
Before feeding the data into the CFAR processor, a square law detector is applied to convert the real and imaginary parts of the radar data into a real-valued square of its power \cite{melebari2015comparison}. 
%According to Equation \ref{equ:range-final}, every frequency bin of the spectrum of the IF signal corresponds to a specific range. Each of these so-called range cells should be analyzed to see whether an object is in the cell. 
The radar signal then runs through a sliding window comprising reference cells, guard cells, and Cell Under Test (CUT). 
%Reference cells are $N$ numbers of cells for detection in the reference window, i.e., the leading window and lagging window. In fact, peaks are not always located in one cell but extend across some range cells. Thus, a small number of guard cells are arranged adjacent to both sides of the CUT, which is tested to detect the presence of the object by comparing the signal level against the noise threshold.
In the estimation of the clutter background level, the utilization of neighboring reference cells surrounding the CUT is justified owing to the spatial-temporal correlation exhibited by radar echoes.
%Due to the spatial-temporal correlation of the radar echo, it is reasonable to employ the surrounding reference cells of the CUT for estimating the clutter background level. 
Then, employing various detectors, the clutter background level of the CUT (referred to as $Z$) can be assessed by utilizing samples within the reference window. The threshold value $T$ is calculated by multiplying $Z$ by a scaling factor $\alpha$. If the value of the CUT (labeled as $Y$) exceeds the threshold value $T$, the comparator will declare that an object is located within the CUT (expressed as $H_1$). Otherwise, there is no object in the current CUT (expressed as $H_0$).

A reasonable selection of processors for the clutter statistical model in the CFAR clutter window can significantly mitigate the issues posed by intricate scenarios, such as uniform clutter, multiple targets, and clutter edges.
%The CFAR processing, whose threshold value is a random variable, is a dynamic process and the threshold value is related to the estimated value of the current clutter background power $Z$. 
As a result of the multitude of techniques available for estimating clutter within the CFAR framework, a variety of CFAR methods have been proposed. These methods can be broadly classified into three types:

\paragraph{Mean-Level Processors}
The first type of processor comprises mean-level estimators, including Cell Averaging CFAR (CA-CFAR) \cite{weiss1982analysis}, Greatest-of CFAR (GO-CFAR) \cite{hansen1980detectability} and Smallest-of CFAR (SO-CFAR) \cite{trunk1978range}. These processors obtain the estimated clutter power level by taking the average value, maximum value and minimum value of the reference cells, respectively. However, the detection mechanism in mean-level processors is not optimized for multiple objects. Other objects within the reference window distort the noise estimation and lead to an increased threshold value, thereby causing potential target detections to be missed.
 
\paragraph{Sorting-Based Processors}
The second category of the processor is exemplified by Ordered Statistics CFAR (OS-CFAR) \cite{rohling1983radar}, which sorts the reference cells in ascending order and selects the $k^ \mathrm{th}$ value as the estimated clutter power level. After that, numerous methods based on ordered statistics have emerged, called OS-like methods, such as Censored Mean Level Detector CFAR (CMLD-CFAR) \cite{rickard1977adaptive}, and Trimmed Mean CFAR (TM-CFAR) \cite{gandhi1988analysis}. However, since the sorting-based processors retain only one ordered reference sample, they rely heavily on prior knowledge about the number and distribution of interference objects. 

\paragraph{Neural Network-Based Processors}
Recently, with the rapid development of machine learning and deep learning, traditional CFAR processors have evolved into neural network-based methods. A representative example of this method is DBSCAN-CFAR \cite{zhao2019robust}, which combines artificial neural network and Density-Based Spatial Clustering of Applications with Noise (DBSCAN) as a clustering algorithm \cite{ester1996density}. 
Unlike conventional methods, DBSCAN-CFAR is able to learn the underlying relationship between normal data and outliers, even in the absence of labeled data or predetermined information regarding the number of clusters.
In simulations involving varying object quantities, shape parameters, and false alarm probabilities, DBSCAN-CFAR exhibits superiority and robustness compared to traditional processors like CA-CFAR, SO-CFAR, GO-CFAR, OS-CFAR, and CMLD-CFAR \cite{zhao2019robust}. However, to achieve the advantages mentioned above, DBSCAN-CFAR suffers from a higher computational burden and time consumption when compared to other processors.

\section{Radar Data Representations}
\label{sec:Radar Data Representations}
In this section, we explore what are radar data representations in autonomous driving and how these representations enhance autonomous driving capabilities.
Firstly, we introduce the generation progress of different radar data representations and their basic characteristics.
Subsequently, for each data representation, we explore the related datasets and representative methods, and discuss the benefits and limitations associated with each representation, thus exploring their capabilities in various perception tasks such as classification, localization, detection, and tracking. 
In addition, we also outline datasets for radar perception in autonomous driving in Table \ref{tab:datasets}.

\begin{figure}[ht]
\begin{center}
\includegraphics[width=1\linewidth]{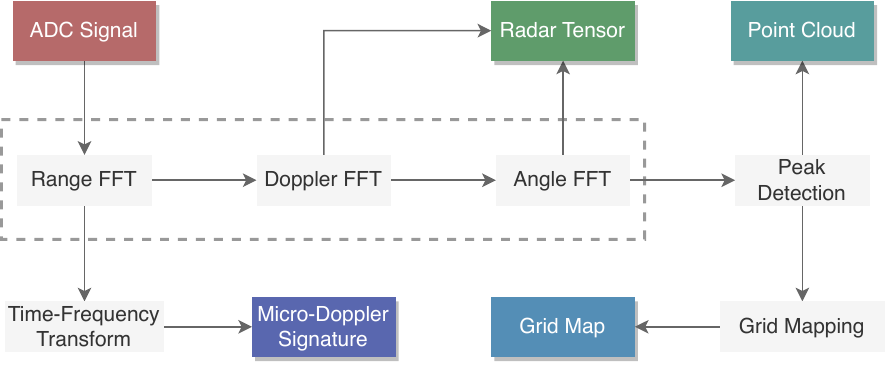}
\end{center}
\caption{Generation progress of five radar data representations (i.e., ADC signal, radar tensor, point cloud, grid map, and micro-Doppler signature).}
\label{fig:radar-signal-processing-chain}
\end{figure}

Fig. \ref{fig:radar-signal-processing-chain} illustrates the generation process of five different radar data representations. 
Initially, the raw output from the radar sensor, known as the \textbf{ADC signal}, which is a digital representation of the analog waveforms received by the radar antenna. The ADC signal is a complex and high-dimensional dataset that contains a mixture of target returns, noise, and clutter, which is hardly interpretable by human observers. 
Thus, researchers utilize 3D FFTs across the sample, chirp, and antenna dimensions to transform the ADC signal into an image-like representation called the \textbf{radar tensor}. The radar tensor organizes the data into a multi-dimensional array where each point corresponds to a specific range, Doppler shift, and angle, making it easier to visualize and analyze the radar returns.
Subsequently, peak detection is employed on the radar tensor to eliminate clutter, resulting in a sparse, point-like representation referred to as the \textbf{point cloud}. Each point in the cloud represents a potential object in the environment, with its coordinates corresponding to the object's range, azimuth, and elevation.
By accumulating point clouds over a specific duration and applying grid mapping methods, the \textbf{grid map} representation is generated for the purpose of identifying static objects. Each cell in the grid map contains information about the occupancy or probability of occupancy of that particular location in the environment.
Moreover, some researchers perform a Time-Frequency transform following the Range-FFT to extract the \textbf{micro-Doppler signature}, which is utilized for recognizing objects characterized by micro-motion features. The micro-Doppler signature is vital for the recognition of objects with discernible micro-motion patterns, enhancing the capabilities of radar systems in various surveillance and identification scenarios.

%In the following contents, we elaborate on the formation process, datasets, methods and discuss advantages and disadvantages associated with each of these representations. Table \ref{tab:datasets} outlines datasets for radar perception in autonomous driving.
%By effective Time-Frequency transform (e.g., Short Time Fourier Transform (STFT) ) on the Range-Time domain of the radar data, the spectrogram containing the MDS is obtained.

\begin{table*}[!h]
\caption{Datasets for radar perception in autonomous driving.}
%\vspace{-3mm}
\scriptsize
\setlength\tabcolsep{5pt} % Decrease this to reduce margins in each cell.
\renewcommand{\arraystretch}{0.8}
\center
%\footnotesize
\begin{tabular*}{\linewidth}{p{1.7cm}<{}p{1.6cm}<{\centering}p{1cm}<{\centering}p{3cm}<{\centering}p{4cm}<{\centering}p{4.8cm}<{}}
\toprule

\bf{Dataset} & \bf{Representations} & \bf{Year} & \bf{Tasks} & \bf{Sensors} & \bf{Scenarios}  \\\midrule

%GUARDIAN \cite{shi2020dataset} & ADC Signal & 2020 & Vital Signs & Radar (TI ADS1298) & Lying down (resting, apnea, tilt up, tilt down)\\\midrule
RaDICaL \cite{lim2021radical} & ADC Signal & 2021 & Object Detection & Radar (TI IWR1443), RGB-D Camera & Indoor (people, static clutter), Roads (urban, rural, highway, various traffic scenarios)\\\midrule
RADIal \cite{rebut2022raw} & ADC Signal, Radar Tensor, Point Cloud & 2021  & Object Detection, Semantic Segmentation & Radar (high-definition), Cameras, LiDAR & Roads (urban, highway, rural) \\\midrule
%\midrule

%\rule{0pt}{-1pt} \rowcolor[HTML]{C0C0C0} & & & & & &\\

CARRADA \cite{ouaknine2021carrada} & Radar Tensor & 2020 & Detection, Semantic Segmentation, Object Tracking & Radar (TI AWR1843), RGB-D Camera, LiDAR & Roads (urban, highway, intersection scenarios)\\\midrule
Zendar \cite{mostajabi2020high} & Radar Tensor, Point Cloud & 2020 & Object Detection, Mapping, Localization & Radar (synthetic aperture), Camera, LiDAR & Roads (diverse urban driving environments) \\\midrule
RADIATE \cite{sheeny2021radiate} & Radar Tensor & 2020 & Object Detection & Radar (Navtech CTS350-X), Camera & Roads (wet, snowy, foggy, rainy, nighttime, urban, highway) \\\midrule
MulRan \cite{kim2020mulran} & Radar Tensor & 2020 & Place Recognition & Radar (Navtech CIR204-H), Cameras, LiDAR & Roads (city, highway, intersection, crosswalks, parks, recreational areas, tunnels, bridges)\\\midrule
Oxford Radar RobotCar \cite{barnes2020oxford} & Radar Tensor, Grid Map & 2020 & Object Detection, Odometry & Radar (Navtech CTS350-X), Camera, LiDAR, GPS, INS & Roads (urban, highway, rural, industrial area, residential area, roundabout, intersection) \\\midrule
SCORP \cite{nowruzi2020deep} & Radar Tensor & 2020 & Semantic Segmentation & Radar (76 GHz), Camera & Roads (parking lot)\\\midrule
CRUW \cite{wang2021rethinking} & Radar Tensor & 2021 & Object Detection & Radar (TI AWR1843, DCA1000), Cameras & Roads (parking, campus, city, highway)  \\\midrule
RADDet \cite{zhang2021raddet} & Radar Tensor & 2021 & Object Detection & Radar (TI AWR1843), Stereo Cameras & Roads (urban, rural, highway, intersections, weather conditions) \\\midrule
Boreas \cite{burnett2023boreas} & Radar Tensor & 2022 & Object Detection, Localization, Odometry & Radar (Navtech CIR304-H), Camera, LiDAR & Roads (highway, rural, urban) \\\midrule
ColoRadar \cite{kramer2022coloradar} & Radar Tensor, Point Cloud & 2022 & Localization & Radar (AWR1843), LiDAR, IMU & Indoor, outdoor environments \\\midrule
K-Radar \cite{paek2022k} & Radar Tensor & 2022 & Object Detection, Object Tracking, SLAM & Radar (RETINA-4ST), Stereo Cameras, LiDAR & Roads (highway, intersection, urban)\\\midrule
OORD \cite{gadd2024oord} & Radar Tensor & 2024 & Place Recognition & Radar (Navtech CTS350-X), GPS/INS & Roads (off-road, difficult terrain, naturalistic environments) \\\midrule
%\midrule

nuScenes \cite{caesar2020nuscenes} & Point Cloud, Grid Map & 2019 & Object Detection, Object Tracking & Radar (Continental ARS408), Camera, LiDAR & Roads (intersection, crosswalk, roundabout, pedestrian crossing)\\\midrule
Astyx \cite{meyer2019automotive} & Point Cloud & 2019 & Object Detection & Radar (Astyx 6455 HiRes), Camera, LiDAR & Roads (highway, urban, rural, parking, roundabout) \\\midrule
SeeingThroughFog \cite{bijelic2020seeing} & Point Cloud & 2020 & Object Detection & Radar (77GHz), Stereo/Gated/FIR Cameras, LiDAR & Adverse road conditions (clear, rainy, snowy, foggy, nighttime, urban, highway, rural, traffic) \\\midrule
%HawkEye \cite{guan2020through} & PC & 2020 & SS & 3D point & Radar () & Roads (highway, intersection, rural, city street lined, narrow mountain pass, coastal road, old town square, tunnel, dirt road, snowy highway) \\\midrule
AIODrive \cite{weng2021all} & Point Cloud & 2020 & Object Detection, Semantic Segmentation, Object Tracking, Depth Estimation & Radar (77GHz), RGB/Stereo Cameras, LiDAR & Roads (highway, residential street, parking)\\\midrule
RadarScenes \cite{schumann2021radarscenes} & Point Cloud & 2021 & Semantic Segmentation, Object Tracking & Radar (77GHz), Documentary Camera & Roads (urban, suburban, rural, highway, tunnel, intersection, roundabout, parking)\\\midrule
Pixset \cite{deziel2021pixset} & Point Cloud & 2021 & Object Detection, Object Tracking & Radar (TI AWR1843), Cameras, LiDARs & Roads (urban, suburban, highway)\\\midrule
%FloW \cite{cheng2021flow} & 2021 & PC & Object Detection & 2D box & camera & waterway \\\midrule
VoD \cite{palffy2022multi} & Point Cloud & 2022 & Object Detection, Object Tracking & Radar (ZF FRGen 21), Stereo Camera, LiDAR & Roads (highway, rural, urban)\\\midrule
TJ4DRadSet \cite{zheng2022tj4dradset} & Point Cloud & 2022 & Object Detection, Object Tracking & Radar (Oculii Eagle), Camera, LiDAR & Roads (intersections, one-way streets) \\\midrule
aiMotive \cite{matuszka2022aimotive} & Point Cloud & 2022 & Object Detection & Radar (77GHz), Camera, LiDAR, GPS, IMU & Roads (highway, urban, rural)\\\midrule
WaterScenes \cite{yao2024waterscenes} & Point Cloud & 2023 & Object Detection, Instance/Semantic/Waterline/ Panoptic Segmentation & Radar (Oculii Eagle), Camera, GPS, IMU & Waterways (river, lake, canal, moat)\\\midrule
%ThermRad \cite{yan2023thermrad} & PC & 2023 & Object Detection, Object Tracking & 3D box, Track ID & Radar (ARS548 RDI, Arbe Phoenix), Camera, LiDAR & Urban, Suburban, Highway, Tunnel, Parking, with weather conditions rainy, cloudy and sunny. (Car, Pedestrain, Cyclist)\\\midrule
NTU4DRadLM \cite{zhang2023ntu4dradlm} & Point Cloud & 2023 & SLAM & Radar (Oculii Eagle), RGB/Thermal Cameras, LiDAR, GPS, IMU & Roads (carpark, garden, campus) \\\midrule
Dual-Radar \cite{zhang2023dual} & Point Cloud & 2023 & Object Detection, Object Tracking & Radar (ARS548 RDI, Arbe Phoenix), Camera, LiDAR & Roads (urban, suburban, highway, tunnel, parking)\\\midrule
MiliPoint \cite{cui2023milipoint} & Point Cloud & 2024 & Activity Recognition & Radar (TI IWR1843), Stereo Camera & Activities (identification, action classification and keypoint estimation) \\\midrule
V2X-Radar \cite{yang2024v2x} & Point Cloud & 2024 & Object Detection & Radar (Oculii Eagle, Arbe Phoenix), Camera, LiDAR & Roads (various weather, time and intersection scenarios) \\\midrule
%V2X-R \cite{huang2024v2x} & Point Cloud & 2024 & Activity Recognition & Radar (TI IWR1843), Camera & Activities (identification, action classification and keypoint estimation) \\\midrule

%millimap \cite{lu2020see} & GM & 2020 & Odometry, OD & Trajectory, 2D box & Radar (Navtech CTS350-X), camera, LiDAR, GPS, INS & Road (vehicle)\\\midrule
%\midrule

Dop-NET \cite{ritchie2020dop} & Micro-Doppler Signature & 2020 & Classification & Radar (Ancortek 24GHz) & Gestures (wave, pinch, click, swipe)\\\midrule
%CI4R \cite{gurbuz2020cross} & Micro-Doppler Signature & 2020 & Classification & Radar (77GHz, 24GHz, Xethru) & Activities (walking, picking, sitting, crawling, kneeling, limping)\\\midrule
Open Radar Datasets \cite{gusland2021open} & Micro-Doppler Signature & 2021 & Classification & Radar (TI AWR2243), Camera, GPS, IMU & Roads (urban, highway, rural)\\\midrule
MCD-Gesture \cite{li2022towards} & Micro-Doppler Signature & 2022 & Classification & Radar (TI AWR1843) & Gestures (push, pull, slide left, slide right, clockwise turning, counterclockwise turning)\\

%\midrule

\bottomrule
\end{tabular*}
\label{tab:datasets}
\end{table*}

\subsection{ADC Signal}

% ADC Signal的原理，特点（雷达的原始数据，只有时间维度信息）；
% 优点：原始数据、信息丰富；
% 缺点：只能用于封闭空间、分类任务。

When an analog signal is sampled and quantized by an ADC, it produces a sequential data stream called the ADC signal. 
As the raw data produced by radar sensors, ADC signals retain all information from the detections, which is highly valuable for deep learning applications. At this stage, the signal lacks spatial coherence among its values, as all information is confined to the time domain \cite{nowruzi2020deep}. To be represented in a more structured format, the ADC signal is typically converted to a 3D Sample-Chirp-Antenna (SCA) tensor, as illustrated in Fig. \ref{fig:ADC}(e).

\subsubsection{Datasets}
With the development of radar technology and increased computing capability, radar data characterized by ADC signal has emerged and garnered widespread attention in recent years.
RaDICaL \cite{lim2021radical} is the first dataset that offers raw ADC signal data collected explicitly for road scenarios in autonomous driving. With access to raw radar measurements, the authors encouraged researchers to design novel processing methods to perform object detection directly or to get downstream data representations.
RADIal \cite{rebut2022raw} is the most comprehensive dataset in terms of radar data representations. In addition to providing ADC signals, the RADIal dataset contains processed data derived from the ADC signals, including radar tensors and point clouds.
Apart from using ADC signals in road scenarios, another significant application is in-cabin vital sign monitoring. However, existing datasets in this area (e.g., \cite{shi2020dataset, gong2021rf, yoo2021radar, chen2022contactless}) primarily focus on indoor scenarios like hospitals or home environments, which differ from in-cabin scenarios in autonomous driving.
Although some studies (e.g., \cite{da2019theoretical, wang2021driver, islam2020non}) investigated monitoring occupant vital signs in cabins, their datasets are unavailable publicly.

\subsubsection{Methods}
The ADC signal representation can be applied for both surrounding traffic perception outside the vehicle and occupant vital sign monitoring inside the vehicle.
Object perception on roads focuses on detecting vehicles and pedestrians using Doppler shifts and other radar features. Vital sign monitoring in a cabin emphasizes extracting and analyzing physiological information from radar ADC signals to monitor occupants.
Since these two situations are handled differently, we classify methods using ADC signals in autonomous driving into road object-based and vital sign-based categories.

\paragraph{Road Object-Based}
% CubeLearn (zhao2023cubelearn) motion recognition
% (stadelmayer2021data) Human Activity Classification
% (stadelmayer2021improved) gait classification
% (arsalan2022spiking) gesture recognition
% ADCNet (yang2023adcnet) 利用ADC进行检测、分割
% T-FFTRadNet (giroux2023t) 利用ADC检测
% Echoes Beyond Points (liu2023echoes) 利用ADC检测
% li2023azimuth 利用ADC增强radar tensor
In recent years, there has been a growing interest in the development of road object perception using ADC signals, as evidenced by research in motion classification (e.g., \cite{stadelmayer2021data, stadelmayer2021improved, arsalan2022spiking, zhao2023cubelearn}) and object detection (e.g., \cite{yang2023adcnet, giroux2023t, liu2023echoes, li2023azimuth}).
These studies encompass two main research directions. 
The first direction involves designing individual end-to-end learnable radar perception models (e.g., ADCNet \cite{yang2023adcnet}, T-FFTRadNet \cite{giroux2023t}, and CubeLearn \cite{zhao2023cubelearn}), enabling object perception on ADC signals. These approaches utilize deep learning on ADC signals as an alternative to traditional signal processing procedures. This solution replaces computationally intensive FFTs and simplifies data flow in embedded implementations, thus significantly reducing the computational requirements.

Another direction of road object-based methods focuses on improving the resolution of radar data, which is essential for estimating object locations and velocities. For example, the ADC Super-Resolution (ADC-SR) model \cite{li2023azimuth} leverages ADC signals to enhance the radar's azimuth resolution, tackling complexities in ADC signals and predicting signals from unseen receivers. Consequently, the hallucinated ADC signals can further refine RAD tensors, enhancing object detection capabilities in autonomous driving.

%\cite{yang2023adcnet, giroux2023t} concentrate on multi-task perception based on ADC signals. 
%Firstly, \cite{yang2023adcnet} proposes an end-to-end radar raw data perception network called ADCNet based on the ADC metrics, combining object detection and semantic segmentation. ADCNet contains a learnable signal processing module, which includes a learnable window and DFT operations of range and Doppler. The learnable signal processing module can implicitly transform the ADC signal into radar tensors with much lower effective signal loss. 
%Secondly, \cite{giroux2023t} proposes a novel end-to-end network called T-FFTRadNet. Like the ADCNet, T-FFTRadNet is also a multi-task perception model, but it first introduces the Vision Transformer \cite{dosovitskiy2020image}, exactly Swin Transformer \cite{liu2021swin} into the radar signal perception. Specifically, T-FFTRadNet first utilizes complex-valued linear layers to introduce the prior Fourier transform \cite{zhao2023cubelearn}. Multiple Swin Transformer blocks are then stacked to extract the hidden features from RAD features after implicit transformation from ADC signals.

\paragraph{Vital Sign-Based}
The vital sign-based methods focus on monitoring the physiological signals of occupants inside the vehicle cabin, including heart rate, respiration rate, and presence detection \cite{diewald2016rf, da2019theoretical, islam2020non, eder2023sparsity, paterniani2023radar}. 
Signal processing techniques, such as spectrogram analysis, Short Time Fourier Transform (STFT) and Wavelet Transform (WT), are employed to extract the vital sign signals from the radar ADC signal \cite{zhang2023overview, paterniani2023radar}. 
Machine learning algorithms, such as statistical classifiers or regression models, are trained using labeled data to estimate vital sign parameters from the extracted features \cite{zhang2023overview}.

Radar systems used for cabin vital sign monitoring are generally designed for close-range operation, such as within a vehicle or confined space. They focus on high-resolution detection of small movements and physiological changes, and use less power and have smaller form factors than road-oriented radar systems. However, vital sign-based methods using machine learning algorithms have limitations in real-time processing, which is crucial in low-latency applications \cite{schires2018vital, le2019radar}.

\subsubsection{Discussion}
% 信息丰富
% 易受干扰
% 数据处理复杂
% 适应室内任务；计算复杂，需要资源多
Given that the ADC signal encompasses raw data derived from object reflections, it inherently carries the richest information about the object. This information can be effectively utilized for diverse tasks involving fine-grained feature recognition, such as activity classification, hand gesture recognition, and vital sign monitoring. 
However, there are certain limitations that should be acknowledged. One limitation is the sensitivity of the radar signals to factors such as electromagnetic interference, occupant positions and obstructing objects, requiring some advanced signal processing methods to extract features.

Another limitation is that ADC signals require a developer kit for data acquisition, which makes data processing challenging. Real-time object perception and vital sign monitoring pose computational challenges, requiring efficient hardware and software solutions.
Moreover, neural networks processing ADC signals are suitable for classification tasks in closed environments, such as determining the presence or absence of objects and distinguishing between different types of objects \cite{stephan2021radar}. However, in open environments, various distracting factors lead to inaccurate object feature identification. 

%Therefore, the use of ADC signals for conducting perception tasks (e.g., object detection, semantic segmentation) by discerning object shape is challenging. 

%Although the radar ADC signal representation shows promising results for object perception on roads and vital sign monitoring in cabins, 

\subsection{Radar Tensor}

% Radar Tensor的原理、特点；
% 优点：数据内容丰富；
% 缺点：存在噪音数据；

\begin{figure}[htbp]
\center
\includegraphics[width=1\linewidth]{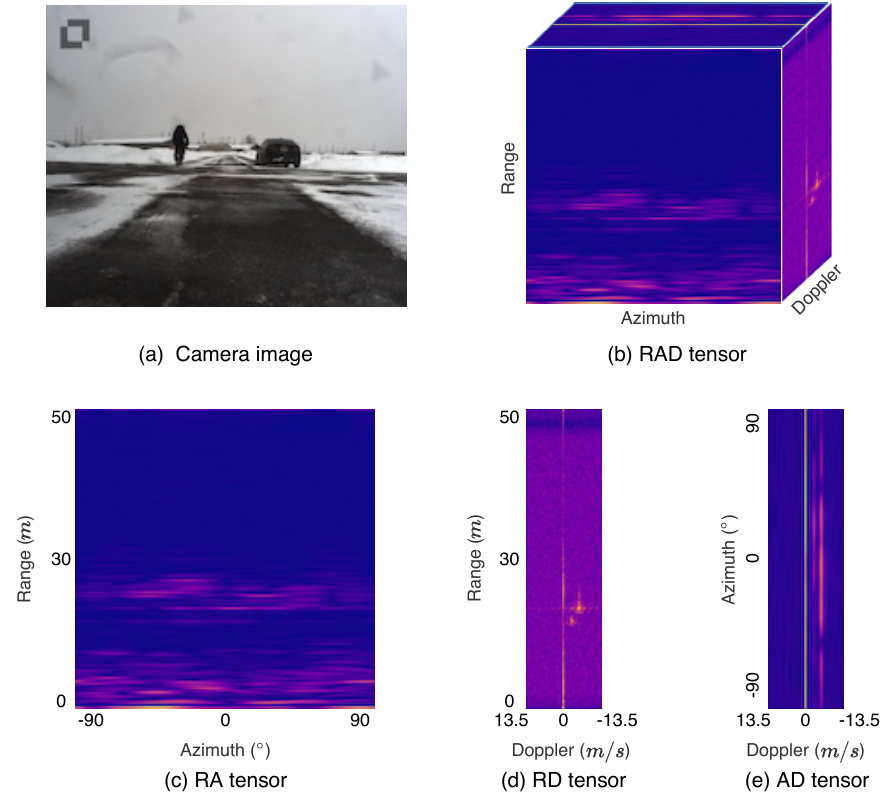}
\caption{Representation of radar tensor. (a) Image from the camera sensor. (b) 3D RAD tensor. (c) 2D RA tensor. (d) 2D RD tensor. (e) 2D AD tensor. Images are generated from the CARRADA \cite{ouaknine2021carrada} dataset.} 
\label{fig:radar-range-azimuth-map}
\end{figure}

As mentioned before, performing 3D FFTs on ADC signals along the sample, chirp, and antenna dimensions yields 3D RAD tensors. 
As is shown in Fig. \ref{fig:radar-range-azimuth-map}, with these three features, two forms of radar tensors are formed: one is in 2D, including the Range-Azimuth (RA) tensor, Range-Doppler (RD) tensor and Azimuth-Doppler (AD) tensor; the other is the whole 3D RAD tensor, with each side consisting of a 2D tensor. 
Specifically, each 2D tensor represents a 2D pseudo-image that describes the spatial pattern of the received echo. The brighter colors within the tensor represent greater reflection amplitude at that location \cite{gao2019experiments}. 
%RA tensor is the spatial representation of the received signals, offering a Bird's Eye View (BEV) of the environment in polar coordinates. 
%To calculate the Direction of Arrival (DoA) map in Cartesian coordinates for detected objects, a Polar to Cartesian transformation is commonly applied to this representation. 
%A series of RD tensors provides a rich temporal feature for dynamic object detection, as dynamic objects exhibit distinct movement patterns compared to stationary objects over time within the RD tensor.
Moreover, in contrast to the 3D radar tensor that lacks height information, the 4D radar tensor expands the data representation by incorporating power measurements in four dimensions: range, Doppler velocity, azimuth angle, and elevation angle. 

\subsubsection{Datasets}
Datasets representing radar data in tensor form can be categorized into three main groups:
\begin{itemize}
  \item \textit{2D tensors:} RADIATE \cite{sheeny2021radiate}, CRUW \cite{wang2021rethinking}, FloW \cite{cheng2021flow}, MulRan \cite{kim2020mulran}, Oxford Radar RobotCar \cite{barnes2020oxford}, SCORP \cite{nowruzi2020deep}, OORD \cite{gadd2024oord};
  \item \textit{3D tensors:} CARRADA \cite{ouaknine2021carrada}, Zendar \cite{mostajabi2020high}, RADDet \cite{zhang2021raddet}, RADIal \cite{rebut2022raw};
  \item \textit{4D tensors:} K-Radar \cite {paek2022k}.
\end{itemize}
The 2D radar tensor dataset has the largest number of all radar tensor datasets. Specifically, RADIATE \cite{sheeny2021radiate}, CRUW \cite{wang2021rethinking}, MulRan \cite{kim2020mulran}, Oxford Radar RobotCar \cite{barnes2020oxford} and SCORP \cite{nowruzi2020deep} are presented in 2D range-azimuth coordinates, representing the Bird's Eye View (BEV) position of objects. The FloW \cite{cheng2021flow} dataset is in 2D range-Doppler coordinates, showcasing the relationship between range and Doppler velocity for each object.
CARRADA \cite{ouaknine2021carrada} is the first dataset that combines synchronized stereo RGB images and 3D radar RAD tensors. It provides annotations with bounding boxes, sparse points, and dense masks for both range-Doppler and range-azimuth representations.
Regarding 4D radar tensors, K-Radar \cite{paek2022k} appears to be the only available dataset, offering comprehensive information on range, Doppler, azimuth, and elevation. 
%However, the size and density of measurements in all four dimensions present challenges, requiring significantly larger memory for storing and processing the data.

\subsubsection{Methods}
Radar tensors are expressed in various forms, with processing algorithms differing depending on tensor dimensions. Therefore, we categorize radar tensor methods into 2D, 3D, and 4D tensor-based methods based on currently available tensor dimensions.

\paragraph{2D Tensor-Based}
2D tensor-based methods typically process 2D input radar data by considering range-Doppler and range-azimuth information.
To conduct object classification or detection on radar tensors, image-based network architectures (e.g., Faster R-CNN \cite{ren2015faster}, ResNet \cite{he2016deep}, U-Net \cite{ronneberger2015u}) are applied or modified, as has been demonstrated in various studies using RA tensors \cite{gao2019experiments, dong2020probabilistic, wang2021rodnet, patel2019deep, huang2022yolo, cozma2021deephybrid} or RD tensors \cite{ng2020range, decourt2022darod, jin2023cross, cozma2021deephybrid}.
For example, 
Gao~\etal~\cite{gao2019experiments} initiated the object detection process by defining a fixed-size bounding box for each detected object on the RA tensor. In addition, they extracted the radar data within the bounding box from multiple RA tensors and arranged this data in the form of radar data cubes. With the temporal information and object movement pattern, the data cubes serve as features for classifying different objects, such as pedestrians, cars and cyclists.
Dong~\etal~\cite{dong2020probabilistic} applied ResNet \cite{he2016deep} architecture on RA tensors for object detection. They also introduced uncertainty estimation for oriented bounding box localization further to enhance the accuracy of the object detection process. 
%Ng~\etal~\cite{ng2020range} performed simulation object detection on RD tensors using a 2D U-Net \cite{ronneberger2015u} as the network architecture.
%Furthermore, Decourt~\etal~\cite{decourt2022darod} proposed a lightweight architecture derived from Faster R-CNN \cite{ren2015faster} and extended the RD tensor-based object detection task to datasets obtained from real scenarios.
In general, 2D tensor-based methods employ a simplified radar data representation, reducing computational complexity. They can exploit correlations between two dimensions to improve detection and localization performance. 
However, these methods only employ part of the available 3D spatial information from the radar. With limited representation, the accuracy of complex scenarios with multiple objects or occlusions may be reduced.

\paragraph{3D Tensor-Based}
3D tensor-based methods process radar data simultaneously using the range, Doppler, and azimuth dimensions, forming a 3D RAD tensor. As such, this approach captures richer spatial information compared to 2D tensor-based methods, thus providing more valuable information to network architectures.
Particular architectures have been developed to process aggregated views of 3D RAD tensors for object detection \cite{major2019vehicle, gao2020ramp, palffy2020cnn, zhang2021raddet, wang2021rodnet} and semantic segmentation \cite{nowruzi2020deep, nowruzi2021polarnet, ouaknine2021multi, dalbah2024transradar}.
Specifically, Major~\etal~\cite{major2019vehicle} first demonstrated the effectiveness of a deep learning-based object detection framework that operated on the RAD tensor and proved that the Doppler dimension helps increase detection performance. 
%They proposed two approaches for handling the RAD tensor. The first method involves compressing along the Doppler dimension and applying mean pooling to obtain the RA model. The second approach involves enlarging the tensor by including two additional 2D tensors, the RD and AD tensors, obtained by compressing along the azimuth angle and range dimensions separately. Feature maps from the three 2D tensors are concatenated together after being resized to the same size. The computational complexity is reduced by processing three 2D tensors instead of directly processing the whole 3D tensor.
Similarly to the approach proposed by Major~\etal~\cite{major2019vehicle}, Gao~\etal~\cite{gao2020ramp} decomposed the RAD tensor into three parts separately before combining them. The primary distinction between the two approaches is that the RA tensor in \cite{gao2020ramp} consists of complex values that aid in recognizing objects using spatial patterns, thus increasing the accuracy of detection.

\paragraph{4D Tensor-Based}
Recently, with the development of 4D radar sensors, methods on 4D radar tensors are emerging.  
K-Radar \cite {paek2022k} compared the performance of 3D Range-Azimuth-Elevation (RAE) tensors with 2D RA tensors, demonstrating the importance of elevation information in radar perception. 
However, this method is still working on the 3D radar tensor without fully using the information of 4D radar tensors.
Later, Enhanced K-Radar\cite{paek2023enhanced} improved K-Radar by introducing a new representation of 4D radar data named 4D Sparse Radar Tensor (4DSRT), which significantly reduces the 4D radar tensor size using offline density reduction. Specifically, 4DSRT retains the top-N\% elements in Cartesian coordinates with the highest power measurements along the XYZ direction. The 3D sparse convolution \cite{liu2015sparse} is then employed to extract the feature maps from the 4DSRT.

%In the pre-processing, the 4DRT is converted from the polar coordinate into the Cartesian coordinate, resulting in a 3DRT-XYZ within the region of interest (RoI). The Doppler dimension is reduced by computing the mean value along the dimension. 
%The 3D sparse convolution \cite{liu2015sparse} is then employed to extracts the feature maps that represent relevant information for bounding box predictions.

Compared to using 2D tensor, 3D tensor-based and 4D tensor-based methods offer a more comprehensive representation of radar data, leveraging the complete spatial information and allowing for precise object detection, segmentation, and localization. These methods can effectively handle complex scenarios involving multiple targets, partial occlusions, and varying azimuth angles. However, the increased dimensions of the tensor result in higher computational complexity and system delays.

\subsubsection{Discussion}
The radar tensor representation combines range, Doppler velocity, azimuth and elevation information into a coherent visual representation. This holistic view of the surrounding objects in the radar's field of view enables efficient perception and environment understanding. With these advantages, radar tensors are widely used in object detection and semantic segmentation tasks through combined labeling with cameras or LiDARs. 

Although radar tensors retain more comprehensive information about an object, they also reserve noise and clutter information, which limit the ability to capture subtle object characteristics or distinguish between similar objects. 
Furthermore, the radar tensor provides a condensed representation of radar data, reducing the overall data size compared to raw ADC signals. Nevertheless, the radar tensor representation still requires considerable memory storage as well as a large bandwidth, especially for the new 4D radar tensors.

\subsection{Point Cloud}

% 点云原理：CFAR之后的数据；表现形式：一系列点的集合；
% 优点：计算量小，过滤噪音数据；
% 缺点：稀疏，重要数据也可能过滤

\begin{figure}[htbp]
\includegraphics[width=1\linewidth]{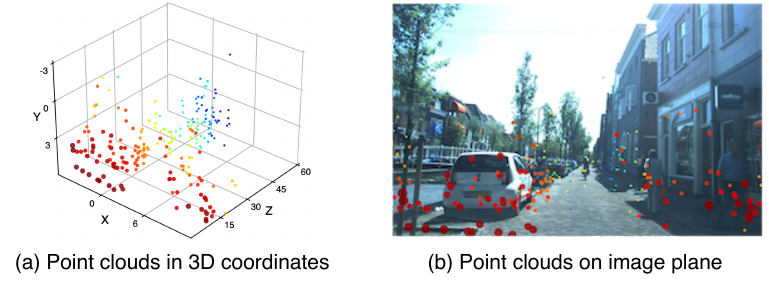}
\caption{Representation of point cloud. (a) Radar point clouds in 3D coordinates. (b) Point clouds projection on 2D image plane. Images are generated from the View-of-Delft \cite{palffy2022multi} dataset.} 
\label{fig:radar-point-cloud}
\end{figure}

By CFAR processing on the radar tensor, data in the format of a group of points, which we refer to as the point cloud, is obtained and visualized in Fig. \ref{fig:radar-point-cloud}(a). The point cloud provides a rough indication of an object's location, yet it cannot accurately describe the outline information of the object, as illustrated in Fig. \ref{fig:radar-point-cloud}(b). Each point contains information about the current reflection, including range, Doppler velocity, azimuth angle, and reflected power.

\subsubsection{Datasets}
%Compared to radar tensors, point clouds serve as a lighter and more intuitive representation of objects. They are also the format of data output from commercial radars. 
Conventional 3D radar sensors generate sparse point clouds, as observed in datasets such as nuScenes \cite{caesar2020nuscenes}, Zender \cite{mostajabi2020high}, SeeingThroughFog \cite{bijelic2020seeing}, HawkEye \cite{guan2020through}, AIODrive \cite{weng2021all}, RADIal \cite{rebut2022raw}, RadarScenes \cite{schumann2021radarscenes}, aiMotive \cite{matuszka2022aimotive} and MiliPoint \cite{cui2023milipoint}.
Recently, 4D radar datasets have emerged, including Astyx \cite{meyer2019deep}, VoD \cite{palffy2022multi}, TJ4DRadSet \cite{zheng2022tj4dradset}, WaterScenes \cite{yao2024waterscenes}, Dual-Radar \cite{zhang2023dual} and V2X-Radar \cite{yang2024v2x}.
While Astyx \cite{meyer2019deep} stands as the first 4D radar point cloud dataset, it is limited by a small data size of only 500 frames. VoD \cite{palffy2022multi} and TJ4DRadSet \cite{zheng2022tj4dradset} datasets represent notable advancements in terms of data categories and data size. Meanwhile, these two datasets also incorporate simultaneous LiDAR data, facilitating comparative analysis between 4D radar point clouds and LiDAR point clouds. 
Interestingly, WaterScenes \cite{yao2024waterscenes} is a 4D radar dataset focused on objects in waterway environments, which demonstrates the robustness of using 4D radar on water surfaces, particularly in adverse lighting and weather conditions.
Dual-Radar \cite{zhang2023dual} incorporates two types of 4D radars, expanding the comparison of different radar performance and further research on practical 4D radar perception algorithms.
V2X-Radar \cite{yang2024v2x} dataset integrates the on-vehicle 4D radar with the roadside 4D radar to achieve the Vehicle-to-Everything (V2X) cooperative perception in autonomous driving.

\subsubsection{Methods}
Point cloud-based radar algorithms take radar point clouds as input and are widely applied for object detection \cite{danzer20192d, tilly2020detection, scheiner2020seeing, xu2021rpfa, xiong2022contrastive}, semantic segmentation \cite{schumann2018semantic, danzer20192d, feng2019point, schumann2018supervised, nobis2021kernel, liu2022deep, kaul2020rss, siddhartha2022panoptic}, tracking \cite{liu2024framework, pan2023moving, deng2023see}, odometry \cite{lu2023efficient, chen2023drio} and scene flow \cite{ding2022self, ding2023hidden, ding2024milliflow}.
Based on various radar point cloud processing techniques, we categorize point cloud-based methods into three groups: clustering-based, point-based, and voxel-based methods.

\paragraph{Clustering-Based}
Clustering-based methods \cite{schumann2017comparison, scheiner2018radar, scheiner2020off} aim to group radar points into clusters using clustering algorithms (e.g., DBSCAN \cite{ester1996density}), and then perform classification based on their spatial attributes. 
While clustering methods may be limited compared to deep learning-based methods, they are still effective in extracting individual objects from radar point clouds. The simplicity of clustering-based methods also reduces memory requirements, making them widely used in radar-based segmentation tasks.
However, clustering-based methods are sensitive to parameter settings, such as the density threshold in DBSCAN. Additionally, these methods may also struggle with cluttered, overlapping, and closely spaced objects.

%However, these methods may struggle with overlapping or closely spaced objects. They are sensitive to parameter settings, such as the density threshold in DBSCAN. Additionally, clustering-based methods may produce fragmented results, especially in cluttered or noisy environments.

\paragraph{Point-Based}
Point-based methods \cite{schumann2018semantic, danzer20192d, feng2019point, tilly2020detection, scheiner2020seeing, pan2023moving, deng2024robust} process radar point clouds at the individual point level, extracting spatial features directly from these points. These methods generally draw inspiration from LiDAR-based algorithms, such as PointNet \cite{qi2017pointnet}, PointNet++ \cite{qi2017pointnet++} and Frustum PointNets \cite{qi2018frustum}. 
These techniques leverage neural networks with convolutional filters to process radar points directly, considering spatial coordinates as integral features in the network structure.
Schumann~\etal~\cite{schumann2018semantic} proposed a structure based on PointNet++ for semantic segmentation on radar point clouds. Specifically, the Multi-Scale Grouping (MSG) module in PointNet++ helps group and generate features for a center point and its neighborhoods. They demonstrated that incorporating the RCS value and compensated Doppler velocity significantly improved classification accuracy.
Inspired by \cite{schumann2018semantic}, Danzer~\etal~\cite{danzer20192d} adopted a two-stage method using PointNets \cite{qi2017pointnet, qi2018frustum} for 2D car detection and segmentation. They considered each point as a proposal and adjusted the proposal size according to the object's prior knowledge. Then, PointNet and Frustum PointNets are used to classify the proposals and each point in the proposal. Finally, the bounding box prediction is executed only for the proposals that are objects. 
%Feng~\etal~\cite{feng2019point} leveraged PointNet and PointNet++ architectures to present the potential of neural networks in radar point cloud segmentation. Their approach primarily emphasizes the tasks of lane, guardrail and road marking segmentation.
Point-based methods are good at accurate localization and classification of objects by considering individual radar points. 
However, they are susceptible to noise and occlusion, which affect feature extraction accuracy.

\paragraph{Voxel-Based}
Voxel-based methods \cite{kim2020grif, xu2021rpfa, nobis2021radar, xiong2023lxl} transform radar point clouds into volumetric representations by discretizing the 3D space into a grid of voxels. Then, 3D convolutions or sparse convolutional neural networks (e.g., VoxelNet \cite{zhou2018voxelnet}, SECOND \cite{yan2018second}) are leveraged to process the voxels to extract features for perception tasks. 
%Pillars are a special form of voxels, spanned over the entire height along the $z$ dimension in a Cartesian coordinate system. Therefore, points are not sorted in separate cells according to their vertical position. 
For example, GRIF Net \cite{kim2020grif} initially converts point clouds into voxels. As point clouds typically exhibit sparsity with numerous empty voxels, GRIF Net utilizes the Sparse Block Network (SBNet) \cite{ren2018sbnet} to convolve only on masked regions, thereby avoiding unnecessary computations on ineffective blank areas. 
In LXL \cite{xiong2023lxl}, the input radar point cloud is initially pillarized as in PointPillars \cite{lang2019pointpillars}. Then, the pillar representation is fed into SECOND to extract multi-level BEV features with spatial and contextual information.
Above all, voxel is an efficient representation for processing dense radar point clouds and is robust to varying point densities and occlusion.
However, voxel-based methods lose fine-grained spatial information due to voxelization and discretization. Moreover, the limited resolution of voxel grids lead to decreased accuracy in object detection, especially for small or distant objects.
\subsubsection{Discussion}
% 去掉了部分噪音

% 信息丢失
Point cloud representation provides an intuitive 3D spatial structure of the surroundings.
Since radar point clouds are generated by filtering techniques, they have advantages in filtering noise and being a lightweight data representation.
However, some potential information within the raw data inevitably be lost \cite{major2019vehicle}. Additionally, some small objects or objects with weak reflections may not be rendered as point clouds.

Currently, most point cloud detection and segmentation algorithms in radar are based on LiDAR algorithms. Point clouds in LiDARs are dense and can describe the outline of surrounding objects. Conversely, radar point clouds are incredibly sparse, posing challenges to developing practical algorithms. To address this issue, point cloud-based algorithms should explore the intrinsic relationships between point clouds, which can provide additional features, including velocities and RCS values.

\subsection{Grid Map}

\begin{figure}[htbp]
\includegraphics[width=1\linewidth]{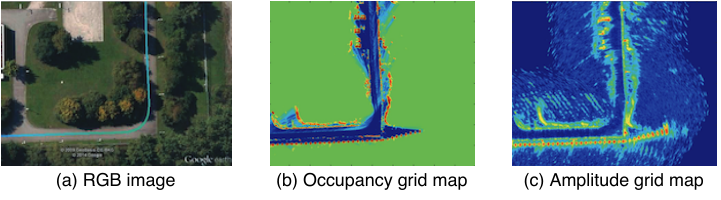}
\caption{Representation of grid map. (a) RGB image from BEV. (b) Occupancy grid map representing obstacles and free-space. (c) Amplitude grid map representing reflected power values. Images are from \cite{werber2015automotive}.} 
\label{fig:radar-grid-map}
\end{figure}

% Grid Map的原理及应用；
% 优点：建图；
% 缺点：需要累计时间，不适用于实时任务
Leveraging multi-frame point clouds, some researchers transform 3D radar reflections into a BEV pseudo-image called the grid map \cite{schumann2018semantic}. While the grid map is presented as aerial views, it is intended to illustrate the complexity and diversity of environments that autonomous vehicles are navigating.
%The measured reflections are integrated over time and inserted at the respective cells in a map. 
There are two primary types of grid maps commonly used in radar-based perception: one is the occupancy-based grid map \cite{werber2015automotive, lombacher2015detection, lombacher2017semantic, degerman20163d, prophet2019semantic, prophet2020semantic, prophet2018adaptions, li2018high, sless2019road, schumann2019scene}, which represents the obstacles and free-space information; the other is the amplitude-based grid map \cite{werber2015automotive, lombacher2015detection}, which displays the reflected power values in that particular area.
As shown in Fig. \ref{fig:radar-grid-map}(b), an occupancy-based grid map has a clear outline, making it more feasible for outline-based detection tasks (e.g., parking spaces). 
An amplitude-based grid map, as indicated in Fig. \ref{fig:radar-grid-map}(c), emphasizes the reflective characteristics of different objects, rendering it more appropriate for rural roads with a few distinct objects.

\subsubsection{Datasets}
There are two main datasets for radar grid map representation.
Based on the nuScenes \cite{caesar2020nuscenes} dataset, Sless \etal\cite{sless2019road} generated a grid map dataset for the semantic segmentation task, containing the actual occupancy state for each cell.
Another grid map dataset is derived from Oxford Radar RobotCar \cite{barnes2020oxford}, a radar dataset using a scanning radar without Doppler information for radar odometry tasks, utilizing the RA tensor as the data representation.
Later, Qian \etal~\cite{qian2021robust} extended this dataset to a grid map representation by creating 3D bounding boxes of vehicles from LiDAR point clouds. To facilitate grid-map object detection, they utilized visual odometry knowledge to synchronize the radar and LiDAR data.

\subsubsection{Methods}
As grid maps are generated from point clouds, clustering algorithms can be used to group radar measurements belonging to the same object. Besides, as the grid map has the same structure as the 2D image, deep learning-based algorithms can be directly implemented for detection and segmentation tasks. 
In general, methods for grid map representations can be categorized into two main types: occupancy-based and amplitude-based methods.

\paragraph{Occupancy-Based}
Occupancy-based methods \cite{werber2015automotive, lombacher2015detection, lombacher2017semantic, degerman20163d, prophet2019semantic, prophet2020semantic, prophet2018adaptions, li2018high, sless2019road, schumann2019scene, popov2023nvradarnet} utilize information such as the presence or absence of objects within a grid cell to make occupancy predictions.
Traditionally, occupancy grid mapping is performed using an Inverse Sensor Model (ISM) and probabilistic methods (e.g., Bayesian filters \cite{li2016effectiveness}) to estimate occupancy probability \cite{prophet2018adaptions, jin2023semantic}. Additionally, Extended Kalman Filter (EKF) approaches take into account both measurement and motion models, modeling the environment as a Gaussian random field and estimating the state of each cell using a recursive Bayesian filter \cite{hussain2022drivable}.
Occupancy-based methods provide a concise environmental representation, simplifying obstacle detection and localization for efficient path planning \cite{jin2021point}.
However, occupancy-based methods struggle to estimate occupancy states for cells with cluttered or overlapping radar returns \cite{srivastav2023radars}. Dynamic objects also pose problems due to their changing nature and intermittent radar measurements.

\paragraph{Amplitude-Based}
Amplitude-based methods (e.g., \cite{werber2015automotive, lombacher2015detection, li2018high, prophet2020cnn}) utilize the measured radar signal reflected intensity information for object detection and classification. 
Representative algorithms group radar measurements with similar amplitudes to identify potential obstacles. Clustering techniques (e.g., DBSCAN, K-means) are often used in this context \cite{li2018high}.
In addition, convolutional neural networks are used as the classifier to distinguish between vehicles and non-vehicles in radar amplitude-based grid maps \cite{lombacher2015detection}.
Amplitude-based methods offer robustness against clutter and noise, as they exploit variations in radar return amplitudes for object detection.
However, amplitude-based representations can be sensitive to sensor noise and uncertainties, as they rely on accurate intensity measurements. The performance of amplitude-based methods is also influenced by factors like object occlusions, multi-path interference, and the radar's limited angular resolution.

\subsubsection{Discussion}
%Occupancy grid mapping has been extensively studied [1], [2] and successfully utilised for a range of tasks including localisation [3], [4] and path-planning []

Radar grid maps provide spatial structure in BEV, offering interpretable and intuitive environmental understanding for geometric localization and obstacle detection. The grid map representation is useful for static objects, as the velocity information is ignored during map construction. By analyzing reflected power values, different object types (e.g., vehicles, pedestrians, buildings) can be distinguished, aiding semantic mapping and scene understanding. 

However, as grid maps are constructed from radar point clouds, point cloud sparsity impacts detection, segmentation, and localization accuracy. This may reduce the accuracy of fine-grained object details or small objects in the grid map. Moreover, closely spaced objects are difficult to distinguish in radar grid maps.

%To accommodate both static and dynamic objects, Schumann~\etal~\cite{schumann2019scene} introduced a two-branch network, separating static and dynamic objects into distinct branches. One branch performs semantic segmentation on grid maps for identifying static objects, while the other branch carries out instance segmentation on point clouds for detecting moving objects. The outputs of the two branches are then merged to generate semantic point clouds. However, this two-branch structure also results in high computational effort and increased memory consumption.

\subsection{Micro-Doppler Signature}

\begin{figure}[htbp]
\begin{center}
\includegraphics[width=1\linewidth]{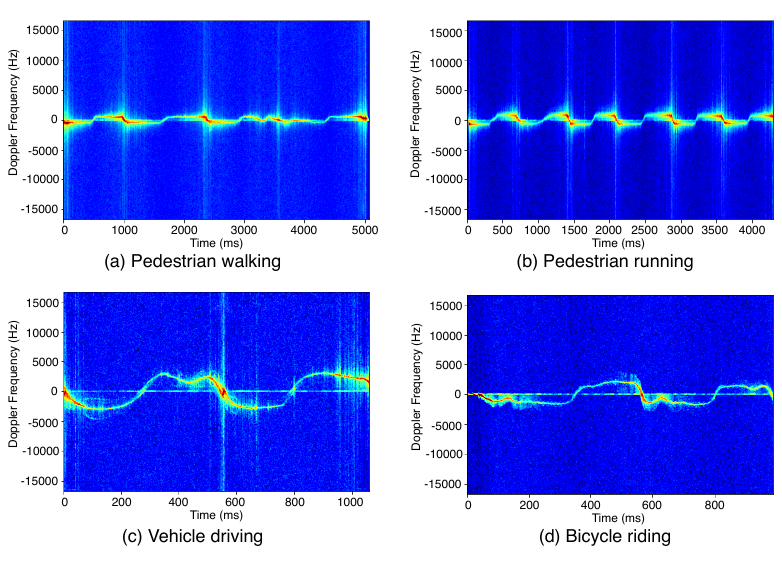}
\end{center}
\vspace{-2mm}
\caption{Representation of micro-Doppler signature. (a) Spectrogram showing a pedestrian walking. (b) Spectrogram showing a pedestrian running. (c) Spectrogram showing a vehicle driving. (d) Spectrogram showing a bicycle riding. Images are generated from the Open Radar Datasets \cite{gusland2021open}.}
\label{fig:radar-micro-Doppler-signature} 
\end{figure}

Micro-Doppler signature refers to the representation of micro-motion, such as rotation and vibration caused by object parts, resulting in a characteristic representation that differs from the Doppler frequency variation.
Micro-Doppler signature is generated by the Time-Frequency transform methods (e.g., STFT, WT) on Range-FFT results.
As illustrated in Fig. \ref{fig:radar-micro-Doppler-signature}, the spectrograms of walking and running pedestrians reveal different features. Specifically, the period of the micro-Doppler for a running pedestrian is shorter than that for a walking pedestrian. Moreover, the spectrograms between different types of objects (e.g., vehicles and bicycles) are unique, making them powerful features for object classification.

\subsubsection{Datasets}
The Open Radar Datasets \cite{gusland2021open} collects different types of outdoor moving targets, such as pedestrian walking and cycling, captured by a stationary radar system. The primary goal of this dataset is to utilize classification techniques for distinguishing between different motion activities.
Apart from outdoor environments, some micro-Doppler signature-based datasets (e.g., Dop-NET \cite{ritchie2020dop}, MCD-Gesture \cite{li2022towards}) for motion recognition (e.g., health monitoring, gesture recognition) are from indoor environments.
These indoor datasets can assist in occupant behavior recognition to some extent, but still differ from behavior classification in autonomous driving due to inconsistent acquisition environments.

%CI4R \cite{gurbuz2020cross} collects 11 different activities and ambulatory gaits. The choice of these activities were motivated by smart environment applications, where monitoring of activities of daily living are required to support health monitoring and gesture recognition.
%Dop-NET \cite{ritchie2020dop} provides four separate hand gestures (wave, pinch, click, swipe), which can be used in human-machine interfaces (HMI) with radar sensor.
%MCD-Gesture \cite{li2022towards} collects data from various domains (i.e. environments, users and locations), and it can be used to develop domain-independent gesture recognition systems based on radar. It provides six predefined gestures (push, pull, slide left, slide right, clockwise turning, counterclockwise turning) and seven other actions as negative samples (lifting right arm, lifting left arm, sitting down, standing up, waving hand, turn around, walking).

\subsubsection{Methods}
Over the past few years, significant research attention has been paid to the classification and recognition of objects' postures and activities through micro-Doppler signature-based methods, particularly in the areas of object classification \cite{gao2019experiments, angelov2018practical}, human activity and gait recognition \cite{gurbuz2019radar, kim2015human, seifert2017new, le2018human}, and human gesture recognition \cite{zhang2019doppler}.
Based on features of the micro-Doppler signature representation, we divide these methods into spectrogram-based methods and cadence velocity diagram-based methods.

\paragraph{Spectrogram-Based}
Spectrograms are a commonly used representation for analyzing micro-Doppler signatures. Spectrogram-based methods simultaneously analyze the micro-Doppler signature in both the time and frequency domains, which has been applied to human motion recognition \cite{kim2015human, le2018human}.
The time-varying gait information, including the speed of arm and leg swing, is effectively encoded within the spectrogram representation. Thus, spectrogram-based methods provide feature localization of micro-Doppler signatures, allowing better identification and classification of different motions.
However, spectrograms suffer from low time-frequency resolution trade-offs, leading to issues in capturing fast-changing Doppler signatures. Besides, spectrograms are also vulnerable to noise interference, which may obscure relevant information.

\paragraph{Cadence Velocity Diagram-Based}
Apart from the spectrogram presenting Doppler information in the time domain, another representation of micro-Doppler signature is the Cadence-Velocity Diagram (CVD) that exists in the frequency domain, which is derived by performing a Fourier transform along the time axis of the spectrogram \cite{bjorklund2012evaluation, seifert2019toward}. 
It provides a metric for capturing the frequency repetition patterns  (``cadence frequencies") over a given period. 
Thus, CVD-based methods provide a valuable measure for periodic changes in body parts at different velocities, such as walking or running \cite{seifert2017new, chen2022attention}. 
However, CVD-based methods become complex when dealing with targets having multiple moving parts or irregular motion patterns. Fine details or rapid changes in motion can be challenging to capture accurately, limiting the ability to distinguish between closely spaced components or rapidly changing motions \cite{ren2023grouped}.

\subsubsection{Discussion}
Micro-Doppler signature recognizes targets by observing minute movement characteristics of objects. 
This representation not only facilitates the differentiation of various object categories (e.g., pedestrians, bicycles, vehicles), but also enables the recognition of intricate object behaviors like gait and gesture recognition. 

However, the transformation to the frequency domain via the Time-Frequency transform suffers from overlapping target components, resulting in suboptimal feature extraction for downstream tasks \cite{stadelmayer2021improved}.
Similarly to ADC signal representation, the micro-Doppler signature is insufficient in delineating an object's spatial coordinates for detection and segmentation tasks, so it is commonly employed for classification tasks. 
Moreover, the micro-Doppler signature representation has limited resolution in range estimation compared to other representations like the radar tensor. This limitation could impact the ability to accurately determine the exact distance of an object, potentially affecting collision avoidance algorithms.

\section{Challenges and Research Directions}\label{sec:Discussion}

In this section, we discuss challenges and potential research directions associated with radar perception in autonomous driving.
When conducting academic research or application development on radar-based perception, the first challenge is what kind of radar product and radar data representation should be selected, as they all have their advantages and application scenarios.
Subsequently, due to the inherent shortcomings of the radar sensor, such as sparsity, noise and limited resolution, we examine how to improve radar perception performance.
Finally, in the trend of multi-sensor fusion with radar as an essential component in autonomous driving, we explore challenges in integrating radar perception with additional sensor modalities, including LiDARs and cameras. This integration aims to enhance the overall accuracy and robustness of perception by leveraging the complementary strengths of each sensor.

\subsection{Which Data Representation Should be Chosen?}
%\subsection{Chosing appreciate }
\begin{table*}[htbp]
\caption{Overview of different radar data representations.}
%\setlength\tabcolsep{5pt}
%\center
%\footnotesize
\begin{tabular*}{1\linewidth}{
p{1.8cm}<{\centering}
p{3.2cm}<{\centering}
p{5.8cm}<{\centering}
p{5cm}<{\centering}
}
\toprule
\centering
\bf{Representations} & \bf{Main Tasks}  & \bf{Advantages} & \bf{Limitations} \\\midrule
ADC Signal & 
\begin{minipage}[t]{1\columnwidth}
\begin{itemize}
\item Classification
\item Object Detection
\item Vital Sign Monitoring

\end{itemize}
\end{minipage} 
& 
\begin{minipage}[t]{0.65\columnwidth}
\begin{itemize}
\item Contains all information from raw radar data 
\item High temporal resolution
\item Fine-grained feature recognition
\end{itemize}
\end{minipage}
&
\begin{minipage}[t]{0.7\columnwidth}
\begin{itemize}
\item Lack of semantic information
\item Complexity in data processing
\item Sensitive to noise
\end{itemize}
\end{minipage}
\\\midrule

Radar Tensor & 
\begin{minipage}[t]{1\columnwidth}
\begin{itemize}
\item Classification
\item Detection
\item Segmentation
\item Localization
\end{itemize}
\end{minipage}  
&
\begin{minipage}[t]{0.65\columnwidth} 
\begin{itemize}
\item Provides spatial structure with range, azimuth and elevation
\item Provides rich motion information about the object
\end{itemize} 
\end{minipage}
& 
\begin{minipage}[t]{0.64\columnwidth}
\begin{itemize}
\item Retains interference information around the object
\item Inconsistent resolution between range and azimuth
%\item Complexity and computational requirements
\end{itemize}
\end{minipage}
\\\midrule

Point Cloud & 
\begin{minipage}[t]{1\columnwidth}
\begin{itemize}
\item Detection
\item Segmentation
\item Tracking
\item Odometry 
\end{itemize}
\end{minipage} 
& 
\begin{minipage}[t]{0.65\columnwidth}
\begin{itemize}
\item Offers 3D spatial representation of the surrounding environment
\item Light-weight representation for objects
\item Filters weak reflected signals
\end{itemize}
\end{minipage} 
&
\begin{minipage}[t]{0.64\columnwidth}
\begin{itemize}
\item Sparse in describing the shape of objects
\item Susceptible to noise and occlusion
\item Loss of potential information after CFAR processing
\end{itemize}
\end{minipage}\\
\midrule

Grid Map & 
\begin{minipage}[t]{1\columnwidth}
\begin{itemize}
\item Mapping
\item Localization
\item Odometry
\item Detection
\end{itemize}
\end{minipage} 
 & 
\begin{minipage}[t]{0.65\columnwidth}
\begin{itemize}
\item Provides structured BEV representation of the surroundings
\item Offers a concise representation focusing on presence and occupancy
\end{itemize}
\end{minipage} 
&
\begin{minipage}[t]{0.64\columnwidth}
\begin{itemize}
\item Difficulty in representing complex and dynamic scenes
\item Suffers from noise, occlusions and resolution
\end{itemize}
\end{minipage}\\
\midrule

Micro-Doppler Signature 
& 
\begin{minipage}[t]{1\columnwidth}
\begin{itemize}
\item Motion Classification 
\item Vital Sign Monitoring
\end{itemize}
\end{minipage} 
& 
\begin{minipage}[t]{0.65\columnwidth}
\begin{itemize}
\item Contains information about object motion
\item Enhances differentiation between objects
%\item Vital sign monitoring
\end{itemize}
\end{minipage} 
&
\begin{minipage}[t]{1\columnwidth}
\begin{itemize}
%\item Precise alignment and synchronization 
%\item sophisticated signal processing techniques may increase the overall system complexity and latency, affecting real-time applications 
\item Limited resolution in range estimation
\item Vulnerable to noise and interference
\end{itemize}
\end{minipage}
\\\bottomrule
\end{tabular*}
\label{tab:comparison}
\end{table*}

As is described in Table \ref{tab:comparison}, different radar data representations have unique advantages and limitations, which should be considered when making informed choices. The choice of data representation depends on the specific requirements of the autonomous driving system, including application scenarios and computational efficiency. 

\subsubsection{Consideration of Application Scenarios}
Autonomous driving vehicles are categorized into six levels of autonomy by the SAE International framework, ranging from Level 0 (no automation) to Level 5 (full automation) \cite{wiseman2022autonomous}. At lower levels of autonomy (Levels 1 and 2), traditional 3D radar sensors are often sufficient for tasks such as adaptive cruise control and lane-keeping assistance, where limited spatial resolution is acceptable. However, as vehicles progress to higher levels of autonomy (Levels 3 and 4), the demand for more precise and robust sensor data increases, making 4D radars a critical component due to their ability to provide denser point clouds, higher resolution, and accurate object tracking. At Level 5, fully autonomous systems rely on a fusion of sensors, including 4D radars, LiDAR, and cameras, to achieve robust perception in complex and dynamic environments.

For applications requiring accurate object detection and tracking, representations like radar tensors or point clouds prove valuable. Radar tensors, such as RAD tensors, offer visualizations of range, Doppler velocity, and azimuth, which can be used to extract features using image-based methods \cite{zhang2021raddet, ouaknine2021multi}. Point cloud representations deliver 3D spatial information aligned with real-world object locations, enabling utilization with point-based network architectures \cite{chamseddine2021ghost, tan20223}. Additionally, point clouds are the output format of conventional radar, benefiting from well-developed product suites and research materials \cite{li2022survey}. 

When considering effective mapping and path planning, grid map representation emerges as a valuable choice. The Grid map offers a structured representation of surroundings, with occupancy grids providing a discretized layout. 
Grids can be classified as occupied or representing free space, aiding vehicle navigation and optimal path planning.

If the research focuses on vehicle cabins, radar representations of ADC signals and micro-Doppler signatures provide powerful insights. By analyzing the characteristics of the ADC signal, such as its amplitude, frequency, and temporal variations, it becomes possible to monitor vital signs.
Additionally, micro-Doppler signatures extract detailed information about human movements, differentiating between vibrations caused by respiratory movements, heartbeats, or even gestures unrelated to vital signs.
Thus, these two representations can be utilized in the cabin for health monitoring, detecting distress, and alerting drivers to emergency services.

% 实时性和计算需求放在一起
\subsubsection{Consideration of Computational Efficiency}
If real-time processing and low computational complexity are critical, computationally efficient representations such as radar tensors and point clouds are preferable. 
The radar tensor represents radar echoes in a 2D or 3D matrix, while the point cloud representation is a set of individual 3D points representing detected objects. As far as computational performance is concerned, some tensor-based methods and point cloud-based methods have already achieved real-time object detection and segmentation \cite{wang2021rodnet, guo2022traffic, isele2021learning, liu2022deep}.

Conversely, ADC signals or 4D radar tensor provide radar data with high temporal resolution, but are typically large and computationally expensive to process in real-time, requiring additional algorithms for signal processing and feature extraction \cite{saponara2019radar, li2023azimuth}. As mentioned in RADIal \cite{rebut2022raw}, the 4D radar tensor at each timestamp amounts to 450 MB, with a processing requirement of 45 GFLOPS for a single elevation, which increases to 495 GFLOPS when all 11 elevation are covered. On the other hand, the point cloud approach requires only 8 GFLOPS to compute the 3D coordinates for a sparse cloud of about 1000 points. In comparison, the processing of a 256-pixel image consumes only 0.4 GFLOPS of computational resources.

Micro-Doppler signatures require additional processing, but their feasibility for real-time depends on specific algorithms and computational resources \cite{li2021human, yao2021radar}.
Grid maps are structured grid-based formats, simplifying subsequent processing and potentially enabling computational efficiency. However, they may not meet real-time requirements if frequent updates are needed for new scenarios \cite{steyer2018grid, hussain2022drivable}.
 
%\subsubsection{Summary} Selecting a data representation for autonomous driving systems requires carefully balancing radar characteristics and application scenarios. It involves considering factors like object detection accuracy, real-time processing capabilities, computational cost, and system scalability. Efficient processing algorithms and hardware acceleration techniques are needed to achieve real-time performance across all representations. 

\subsection{How to Improve Radar Perception Performance?}
The constraints inherent in radar sensors and the limitations associated with radar data processing algorithms present challenges to the performance of radar perception. Improving radar perception performance is crucial for autonomous driving, enabling more accurate decisions and predictions. In the following, we explore four areas: creating high-quality datasets, dealing with inherent limitations, improving data processing algorithms, and employing 4D radar sensors. An overview of challenges and research directions in these four areas is described in Table \ref{tab:challenges}.

\begin{table*}[htbp]
\caption{Overview of challenges and research directions on improving radar perception performance.}
\setlength\tabcolsep{5pt}
%\center
\footnotesize
\begin{tabular*}{1\linewidth}{
p{5.1cm}<{}
p{4.9cm}<{\centering}
p{8.2cm}<{\centering}
}
\toprule
\centering
 \bf{Topic} & \bf{Challenges} & \bf{Research Directions} \\\midrule
Creating High-Quality Datasets & 
\begin{minipage}[t]{0.3\textwidth}
\begin{itemize}
  \item Calibration
  \item Annotation
\end{itemize}
\end{minipage}
&
\begin{minipage}[t]{0.55\textwidth}
\begin{itemize}
  \item Develop adaptive and self-calibration algorithms
  \item Design real-time and dynamic calibration techniques
  \item Explore auto-labeling techniques
\end{itemize}
\end{minipage}
%\\\cmidrule{2-4}
\\\midrule

Dealing with Inherent Limitations
&
\begin{minipage}[t]{0.3\textwidth} 
\begin{itemize}
  \item Sparsity and limited resolution
  \item Noise and clutter
\end{itemize} 
\end{minipage}
& 
\begin{minipage}[t]{0.45\textwidth}
\begin{itemize}
  \item Explore the distribution of radar output
  \item Construct features based on physical properties
  \item Model inherent uncertainty
\end{itemize}
\end{minipage}
%\\\cmidrule{2-4}
\\\midrule

Improving Data Processing Algorithms &
\begin{minipage}[t]{0.3\textwidth}
\begin{itemize}
  \item Parameter estimation
  \item Feature extraction
\end{itemize}
\end{minipage} 
&
\begin{minipage}[t]{0.45\textwidth}
\begin{itemize}
  \item Use neural networks instead of traditional FFTs
  \item Apply graph neural network to extract features
  \item Utilize occupancy prediction for scene understanding 
\end{itemize}
\end{minipage}\\
\midrule

Employing 4D Radar Sensors &
\begin{minipage}[t]{0.3\textwidth}
\begin{itemize}
  \item Large size 
  \item Perception algorithms
\end{itemize}
\end{minipage} 
&
\begin{minipage}[t]{0.45\textwidth}
\begin{itemize}
  \item Extract core values from tensor dimensions
  \item Develop efficient 3D object perception models
  \item Design specific architectures on dense information
\end{itemize}
\end{minipage}
\\\bottomrule

\end{tabular*}
\vspace{1mm}
\label{tab:challenges}
\end{table*}

% 构建高质量数据集（标定和标注）
\subsubsection{Creating High-Quality Datasets}
A high-quality dataset provides more accurate ground truth information, enabling the algorithm to learn the correct associations and features required for accurate object classification, detection, tracking, and localization. 
However, creating a high-quality dataset is challenging for radar, primarily in terms of calibration and annotation.

\paragraph{Calibration}
Ensuring accurate calibration for radar sensors is a prerequisite for radar perception. 
The triangular corner reflector is a common target for radar calibration. When radar waves encounter the corner reflector, they bounce off each surface and converge at a point, forming a strong reflection that the radar sensor can easily measure \cite{michelson1995depolarizing}.
While significant progress has been made in radar calibration techniques (e.g., \cite{domhof2021joint, cui20213d, zhang20223dradar2thermalcalib}), there are still challenges to address and potential research directions to explore.
%Proper alignment, timing synchronization, and correction of biases or distortions in the radar measurements help generate more reliable and accurate perception results.

Designing calibration methods adapted to varying environmental conditions is essential. Research efforts should focus on developing adaptive and self-calibration algorithms that account for changes in clutter, interference, and propagation conditions \cite{petrov2021auto}.
Developing advanced algorithms that handle complex system configurations is a promising research direction. Techniques such as machine learning, optimization, and statistical approaches can be explored to improve calibration accuracy and efficiency \cite{li2023globally}.
In addition, dynamic radar calibration during driving is challenging. Developing dynamic calibration techniques, such as the Gaussian Mixture Model (GMM), that adapt to changing systems and environments enhances the operational reliability of radar systems \cite{li2023globally}.
Addressing these challenges and exploring these research directions can lead to improved radar calibration techniques, enabling more accurate and reliable radar measurements in autonomous driving.

\paragraph{Annotation}
In another aspect, high-quality datasets with accurate and rich annotations are crucial for training and validating perception algorithms. 
However, the process of data annotation is labor-intensive and time-consuming. This is particularly true when dealing with radar data, as the inherent representation of objects within this data modality does not fully reveal their physical forms.
Data describing objects with ground truth information, such as position, velocity, reflected power, and class, should also be annotated to enable training models that generalize well to real-world conditions.

The exploration of auto-labeling techniques for radar data presents a promising avenue to mitigate the burden of manual data labeling. In practice, radar data labels can be derived by leveraging corresponding ground truth from camera images and the extrinsic matrix of radar and camera sensors \cite{domhof2021joint, yao2024waterscenes}. However, the efficacy of this labeling approach for radar data remains an issue, given that radar targets may not consistently align with the ground truth depicted in images.
Thus, despite the potential benefits associated with auto-labeling radar data, effectively filtering extraneous data around objects remains an ongoing challenge.

%low resolution, clutter, trade-off
\subsubsection{Dealing with Inherent Limitations}
While radar sensors offer significant advantages for perception in autonomous driving, they also have certain limitations. 
Factors such as adverse weather conditions, interference, and the trade-offs between range, resolution, and field of view are significant considerations that impact the reliability and effectiveness of radar systems. Understanding these limitations is crucial for designing appropriate strategies to create a robust and comprehensive perception system.

%数据稀疏（稀疏、分辨率低）
\paragraph{Sparsity and Limited Resolution}
%Addressing the sparsity and ambiguity of radar data presents unique and complex challenges.
The sparse nature of radar data provides less detailed information about object shape and fine structure than LiDAR and camera data.
Limited resolution hinders the radar system's ability to distinguish between closely spaced targets, leading to potential confusion and misinterpretation. 
Consequently, radar sensors struggle to provide precise localization or detailed shape information, making it challenging to classify and discern fine-grained object features.

The frequency of radar systems plays a critical role in determining data representation parameters, particularly resolution and the ability to discriminate between surfaces and objects with varying textures or roughness. Modern automotive radars, which operate at higher frequencies such as 77 GHz or 79 GHz, benefit from increased bandwidth, leading to improved range resolution. This allows for better differentiation between closely spaced objects and enhances the detection of subtle variations in distance. 
In addition to spatial resolution, higher frequencies enhance the radar's ability to detect variations in surface textures and material properties, aiding in object classification and recognition. However, these advantages come with trade-offs, such as increased signal attenuation due to atmospheric absorption and reduced penetration through certain materials, which must be carefully considered in system design.

While combining multiple frames enhances accuracy, it can also introduce system delays \cite{schumann2018semantic, nobis2019deep, nabati2021centerfusion, kim2020grif, long2021radar, palffy2022multi, harley2023simple}.
Column or pillar expansion methods (e.g., \cite{nobis2019deep, nabati2021centerfusion, li2020feature}) complement the lack of measurement in the vertical direction, but only alleviate ambiguity to a limited extent.
Exploring radar output distribution (e.g., Gaussian distribution) based on the fundamental principles of radar presents a valuable research opportunity \cite{stacker2022fusion}.
Moreover, existing methods (e.g., PointNet \cite{qi2017pointnet}, PointNet++ \cite{qi2017pointnet++}) designed for processing dense LiDAR point clouds are limited for handling sparse radar point clouds \cite{wang2018sgpn}. Therefore, developing specialized architectures to address the challenges of sparse radar detection presents a promising research direction.
For example, Liu~\etal~\cite{liu2022deep} leveraged cosine similarity loss and normalized inner product loss as part of the training process for sparse radar detection points, enhancing the performance of point-wise center shift vector-guided clustering.

While conventional radar systems with a limited number of antennas may exhibit constrained angular resolution, this limitation can be effectively addressed through Synthetic Aperture Radar (SAR) techniques. By exploiting the motion of the vehicle, SAR can synthesize a large virtual aperture through advanced signal processing techniques, thereby enhancing angular resolution. This approach has been extensively studied in recent works \cite{tagliaferri2024cooperative, tagliaferri2021cooperative, manzoni2024automotive, manzoni2022motion, tagliaferri2021navigation, tebaldini2022sensing, rizzi2022multi}, demonstrating its feasibility and effectiveness in automotive radar applications. For instance, \cite{tagliaferri2024cooperative} and \cite{tagliaferri2021cooperative} explored cooperative imaging in multi-sensor networks, particularly addressing phase synchronization and data processing to achieve high-resolution imaging. \cite{manzoni2024automotive, manzoni2022motion, tagliaferri2021navigation} analyzed the potential and challenges of automotive SAR imaging, proposing motion estimation and compensation techniques and navigation-aided imaging methods to mitigate the impact of vehicle motion on imaging quality. \cite{tebaldini2022sensing} and  \cite{rizzi2022multi} analyzed the impact of urban environments on SAR imaging, proposing multi-beam SAR imaging techniques to address multi-path effects and occlusion issues in complex urban settings.

Given SAR's ability to achieve spatial resolution comparable to LiDAR, the representation and storage of SAR data require careful consideration. High-resolution SAR data can be represented as point clouds, range-Doppler maps, or grid maps, with point clouds being particularly suitable for applications like autonomous driving due to their compatibility with LiDAR-based systems. However, the increased data volume necessitates efficient storage solutions, such as compression algorithms and hierarchical data structures. The choice of output format depends on the specific application, with point clouds being preferred for environment mapping and range-Doppler maps for tasks like target detection.

% 干扰多（杂波、多径）
\paragraph{Noise and Clutter}
Radar sensors are susceptible to noise, resulting in false detections and inaccurate object positions. Noise arises from various sources, including electronic components within the radar system, thermal noise, atmospheric disturbances, and electromagnetic interference. Clutter refers to unwanted radar returns caused by environmental factors such as ground reflections, vegetation, or other structures. 

The signal received by the in-cabin radar sensor not only contains information about the occupant's vital signs, but also includes unwanted motion induced by engine vibrations \cite{da2019theoretical}. Consequently, monitoring vital signs in a non-stationary environment using a radar sensor becomes rather challenging.
Traditional techniques (e.g., \cite{wang2021cfar, zhang2018support, jin2019automotive, alhumaidi2021interference}) applied to reduce noise in automotive radars usually draw upon CFAR and peak detection algorithms. However, these methods have shown limitations in terms of adaptability when faced with varying conditions and unpredictable noise types. 

For noise removal in radar tensors, researchers draw inspiration from principles and techniques used in image denoising and restoration \cite{schussler2022deep, dubey2020region, ristea2020fully, fuchs2020automotive, chen2021dnn, chen2022two, de2020deep}. By considering radar tensors as images, various techniques, including convolutional neural networks, restoration filters, and wavelet-based methods, can be applied to reduce noise and enhance the quality of these tensors \cite{ilesanmi2021methods, rasti2021image}.
For mitigating noise in point clouds, removing radar detections that fall outside boundaries defined by ground truth information from cameras or LiDARs is an approach \cite{nobis2019deep, cheng2021new}.
Constructing features based on physical properties can provide the necessary guidance for models to differentiate targets from clutter. This involves examining point density, spatial arrangement, and other statistical properties to help distinguish signal from noise.
However, a unique challenge presented by radar point clouds is their inherent sparsity and inaccuracy.  Developing techniques to account for this uncertainty could allow for more robust noise removal and aid in distinguishing objects from noise.

\subsubsection{Improving Data Processing Algorithms}

Processing raw radar data to extract meaningful object information is critical for perception accuracy. 
Advanced signal processing techniques suppress clutter and noise in radar measurements, including adaptive beamforming and spatial filtering \cite{nosrati2020multi, kumari2021adaptive}.
However, data processing using these signal processing techniques remains challenging for radar perception. Utilizing machine learning and deep learning algorithms can also improve perception and object recognition capabilities. We focus on discussing two aspects of algorithm improvement: parameter estimation and feature extraction.

\paragraph{Parameter Estimation}
Radar signals often exhibit complex patterns and non-linear dependencies. Traditional FFT-based methods only capture part of the relevant information required for parameter estimation \cite{hakobyan2019high, wang2021rodnet}.
One potential research opportunity is employing neural networks to extract radar parameters instead of traditional FFT operations. This approach is valuable to reduce computational requirements, while simplifying the data flow within embedded implementations \cite{huang2018deep, wang2021rodnet, ma2022deep}. An example can be found in RODNet \cite{wang2021rodnet}, where FFT operations are selectively applied in the sample and antenna dimensions, while the chirp dimension is preserved to form the range-azimuth-chirp tensor. 
Then, a neural network is employed to process the chirp dimension for extracting relevant Doppler features. 
This end-to-end learning approach allows the neural network to learn relevant features and patterns from the radar data, potentially improving the accuracy and efficiency of parameter extraction.

\paragraph{Feature Extraction}
Graph Neural Networks (GNNs) \cite{scarselli2008graph} present a promising research direction, enabling operations on graph data to effectively capture relationships between elements within complex structured data (e.g., point clouds, images) \cite{scarselli2008graph}.
Radar data can also be represented as a graph, where each graph node represents a point from point clouds or a pixel from radar tensors, and the edges capture the spatial or temporal relationships between them. 
Then, Graph Convolutional Networks (GCNs) employ graph convolutional layers to update node embeddings by aggregating and transforming information from neighboring nodes. These layers effectively capture spatial dependencies between objects, facilitating feature learning in graph-based radar data representations.
GNN-based methods applied to radar point clouds (e.g., Radar-PointGNN \cite{svenningsson2021radar}, RadarGNN \cite{fent2023radargnn}) or radar tensors (e.g., GTR-Net \cite{meyer2021graph}) demonstrate graph representations' effectiveness in capturing contextual and spatial information, thereby improving the overall performance.

%To dig deeper into the relationship between sparse radar point clouds, GNN  is a promising research direction in which each point is considered as a node, and edges are the relationship between the points. GNNs are deep learning models designed to operate on graph data, allowing them to effectively capture relationships and dependencies among elements in complex structured data, such as point clouds, graphs, or networks.

%In Radar-PointGNN \cite{svenningsson2021radar}, GNN adopted for feature extraction of radar point clouds demonstrates that the graph representation produces more effective object proposals than other point cloud encoders by mapping radar point clouds to contextual representations. 
%RadarGNN \cite{fent2023radargnn} indicates that GNNs can operate on unstructured and unordered data, obtaining both point features and point-pair features embedded in the edges of the graph. Thus, compared to voxelization operations, GNN eliminates the information loss from the sparse radar point clouds.
%GNN also shows its advantages in detection from RA tensors. The Graph Tensor Radar Network (GTR-Net) \cite{meyer2021graph} architecture utilizes graph convolutional operations to aggregate information across the point cloud nodes. The process involves weighting the features of connected nodes based on their respective edge weights. In this way, it improves the defective sparse points by aggregating relevant information and thus leads to better performance.

Recently, occupancy prediction (e.g., SurroundOcc \cite{wei2023surroundocc}, OpenOcc \cite{tong2023scene}, Occ3D \cite{tian2023occ3d}, TPVFormer \cite{huang2023tri}) using cameras and LiDARs has been a hot research topic in autonomous driving. Compared to 3D bounding boxes, 3D occupancy can not only describe target objects, but also indicate background state and capture fine-grained details \cite{tong2023scene}.
Radar occupancy prediction presents a valuable research direction for radar perception in autonomous driving. By analyzing occupancy patterns in radar data, the structure of the scene, including road boundaries as well as obstacles, can be inferred. 
Potential research could explore using contextual semantics to enhance radar occupancy prediction accuracy, enabling more comprehensive environment understanding.

%However, radar-based occupancy prediction faces challenges in constructing accurate and up-to-date grids that capture object presence or absence.

%\subsubsection{Vital}
%\cite{islam2020non} Chirp parameters and signal processing steps were developed to extract phase information for signals reflected from tiny movement of a subject's chest surface. Beam steering techniques were used to isolate the respiratory signatures for individual subjects from radar signals reflected simultaneously from multiple subjects.
%\cite{liu2023echoes}

% 利用4D radar
\subsubsection{Employing 4D Radar Sensors} 
Radar sensors have made significant progress, transitioning from 3D to 4D capabilities, including enhanced resolution and elevation measurement capabilities. As a result, more and more research is focused on the 4D radar, with growing numbers of radar datasets and algorithms proposed recently. Given these advancements, employing 4D radar sensors is a promising and potential research direction for radar perception in the broader domain of autonomous driving. However, 4D radar is still challenging in terms of large data size as well as perception algorithms.

%4D radars offer superior spatial resolution and elevation measurement, generating denser radar data. More and more 4D radar datasets and algorithms have been proposed in recent years. 
\paragraph{Large Size}
The spatial distribution of objects is effectively captured and represented within 4D radar datasets, such as VoD \cite{palffy2022multi}, TJ4DRadSet \cite{zheng2022tj4dradset}, K-Radar \cite {paek2022k}, WaterScenes \cite{yao2024waterscenes}, and Dual-Radar \cite{zhang2023dual}. 
However, compared to 3D radar sensors, 4D radars produce prohibitively large data outputs.
For example, in the K-Radar \cite {paek2022k} dataset, the size of 4D radar tensor only in the forward direction amounts to 12TB, while the size of the synchronously acquired 360-degree LiDAR point cloud data is only 0.6TB.
Thus, methods of size reduction are worth consideration. Preliminary investigations involve developing Enhanced K-Radar \cite{paek2023enhanced}, which extracts higher value by sampling different tensor dimensions. This innovative approach shows potential for enhancing training speed and reducing memory requirements.

\paragraph{Perception Algorithms}
Emerging perception algorithms applied to 4D radar datasets also indicate that the 4D radar data significantly enhances radar perception capabilities \cite{xu2021rpfa, palffy2022multi, liu2023smurf, yan2023mvfan, tan20223}.
%For example, Palffy~\etal~\cite{palffy2022multi} feed 4D radar data with five features (spatial coordinates, reflectivity, and Doppler velocity) into PointPillars \cite{lang2019pointpillars} and pointed out that the additional elevation data increases object detection performance (from 31.9\% to 38.0\% in mAP) in their VoD \cite{palffy2022multi} dataset.
%RadarMFNet \cite{tan20223} adopts a modified PointPillars variant to effectively capture spatiotemporal features by integrating information from the current scan with consecutive scans. However, they don't sufficiently tackle the adverse effects of noise points inherent in the 4D radar point cloud data.
%To overcome the limitations posed by sparse and noisy data, SMURF \cite{liu2023smurf} introduces a novel approach based on Kernel Density Estimation (KDE) \cite{parzen1962estimation} for extracting density features from the 4D radar point cloud. 
With various architectures (e.g., point-based, voxel-based, pillar-based) proposed, Palmer~\etal~\cite{palmer2023reviewing} conducted a comprehensive analysis of detection performance achieved by existing models on 4D radar datasets, including VoD \cite{palffy2022multi} and Astyx \cite{meyer2019deep}. They evaluated the performance of Voxel R-CNN \cite{deng2021voxel}, SECOND \cite{yan2018second}, PointRCNN \cite{shi2019pointrcnn}, and PV-RCNN \cite{shi2020pv} through cross-model validation and cross-dataset validation experiments. Numerical results showed no clear best model, with the performance varying by object class and distance. 
Further research is expected to develop efficient 3D object detection models based on 4D radar data, utilizing rich data and feature extraction to address challenges.

\subsection{How Can Data Fusion Be Applied to Different Radar Representations?}
%\subsection{Data Fusion for Radar Representation?}
While radar sensors are good at detecting and ranging objects, they provide limited information about object characteristics like color, texture, and fine-grained visual details. Integrating radars with complementary sensors, such as cameras, LiDARs and ultrasonic sensors, overcomes this limitation and provides comprehensive perception solutions \cite{yao2023radar, feng2020deep, yao2023radarperception, yao2024towards, wiseman2018ancillary}. Researchers can leverage the complementary strengths of each modality to achieve a balance between performance, cost, and robustness. Sensor fusion techniques further enhance system reliability by combining range and velocity data from radar with high-resolution spatial data from other sensors.

When fusing radar from multiple sensors, the choice of representation depends on the specific application requirements, the type of radar systems involved, and the desired level of information granularity. Common approaches to multi-radar fusion can be categorized into three levels: data-level fusion, feature-level fusion, and object-level fusion \cite{yao2023radar}. Data-level fusion involves combining the unprocessed radar signals from multiple radars. This approach preserves the maximum amount of information, as it avoids any loss of detail during intermediate processing steps. Feature-level fusion extracts meaningful features from the raw data of each radar system, such as target range, velocity, Doppler shift, and angular position. These features are then combined to form a unified representation of the detected targets. This approach strikes a balance between computational efficiency and information preservation, making it a popular choice for many practical applications. Object-level fusion involves processing the raw data from each radar system independently to make local decisions (e.g., object detection, classification, or tracking) and then combining these decisions to produce a final result. Common methods for object-level fusion include majority voting, weighted averaging, and Bayesian inference.
However, radar data fusion also faces some challenges, particularly in heterogeneous data association and in challenging condition processing. An overview of these challenges and research directions is summarized in Table \ref{tab:challenges-fusion}.

\begin{table*}[htbp]
\caption{Overview of challenges and research directions on radar data fusion.}
\setlength\tabcolsep{5pt}
%\center
\footnotesize
\begin{tabular*}{1\linewidth}{
p{4cm}<{}
p{5.3cm}<{\centering}
p{5.7cm}<{\centering}
}
\toprule
\centering
\bf{Topic} & \bf{Challenges} & \bf{Research Directions} \\\midrule
Associating Heterogeneous Data & 
\begin{minipage}[t]{0.29\textwidth}
\begin{itemize}
  \item Projection-based or geometric-based associations result in ineffective alignment
  \item Sensitive to occluded radar returns and background clutter
\end{itemize}
\end{minipage}
&
\begin{minipage}[t]{0.43\textwidth}
\begin{itemize}
  \item Utilize BEV features to enhance spatial understanding
  \item Apply attention mechanisms to mitigate background clutter
  \item Employ joint probabilistic data association methods to estimate the likelihood of associations
\end{itemize}
\end{minipage}
%\\\cmidrule{2-4}
\\\midrule

Handling Challenging Conditions
&
\begin{minipage}[t]{0.29\textwidth} 
\begin{itemize}
\item Sensitive to adverse weather and object occlusions
\item Uncertainty estimation of different sensor measurements
\end{itemize} 
\end{minipage}
& 
\begin{minipage}[t]{0.43\textwidth}
\begin{itemize}
  \item Develop adaptive fusion strategies to adjust fusion parameters based on environmental conditions
  \item Employ probabilistic fusion techniques to model and propagate uncertainties 
\end{itemize}
\end{minipage}
\\\bottomrule

\end{tabular*}
\vspace{1mm}
\label{tab:challenges-fusion}
\end{table*}

\subsubsection{Associating Heterogeneous Data}

% 数据关联问题：
% 直接投影是有问题的，
% 在于 1：不是一一对应的关系，且雷达点不准；
% 潜在：BEV

A significant challenge is the ambiguity in associating radar data with other modalities, as they are heterogeneous.
For radar-camera fusion, existing approaches involve projecting radar data onto the image plane and subsequently establishing their correspondence via a calibration matrix \cite{chadwick2019distant, meyer2019deep, nobis2019deep, bai2021robust, emami2021long}. For radar-LiDAR fusion, existing methods exploit geometric information to associate radar and LiDAR point clouds by comparing the position, velocity, and shape of the detected objects to determine if they correspond to the same physical object \cite{wang2022multi}.
However, projection-based or geometric-based associations result in ineffective alignment with object centers. As previously stated, radar data exhibits sparsity, inaccuracy, and noise, thereby causing inadequate associations in both object-level and data-level fusion scenarios.

Therefore, associating radar data with other modalities is a critical but challenging question. In our opinion, the incorporation of BEV features, transformer architectures, and attention mechanisms are valuable research directions that can significantly improve fusion performance.
BEV provides a top-down perspective view of the vehicle's surroundings, thereby enhancing spatial understanding.
Using BEV transformations and transformer architectures, radar-camera fusion methods (e.g., RCBEV \cite{zhou2023bridging}, CRAFT \cite{kim2023craft}, CRN \cite{kim2023crn}, RCFusion \cite{zheng2023rcfusion}, LXL \cite{xiong2023lxl}) or radar-LiDAR fusion methods (e.g., Bi-LRFusion \cite{wang2023bi}, BEVGuide \cite{man2023bev}, ACF-Net \cite{tian2023acf}, ST-MVDNet++ \cite{li2023st}) have proven to deliver impressive performance.

Moreover, attention mechanisms have been successfully utilized to deal with complex contextual relationships, which makes them promising for data fusion tasks \cite{yao2023radar}. Attention mechanisms play a crucial role in determining the relevance of different sensor modalities at different spatial or temporal locations. In the case of data association, the attention mechanism filters occluded radar returns and mitigates background clutter, which facilitates the alignment and matching of objects or features across the different modalities \cite{long2021radar, kim2023craft}. Thus, attention-based association with adaptive thresholds emerges as a prospective approach for establishing a connection between radar and other modalities.
%Investigating attention mechanisms for radar-camera and radar-LiDAR fusion can help the system focus on relevant sensor inputs while suppressing noise and irrelevant information. 
Furthermore, the Joint Probabilistic Data Association (JPDA) method is a potential research direction that leverages Bayesian filtering techniques to estimate the likelihood of associations based on the joint probability distribution of radar and other modalities \cite{liu2021robust}. By considering uncertainties associated with the measurements and estimates, JDPA enables robust data association.

%Attention mechanisms have been successfully applied to deal with complex contextual relationships, which makes them promising for data fusion tasks as well. 
% 融合架构：BEV, attention, uncertainty
%\subsubsection{Design Effective Architectures}
%\paragraph{Network Architecture}
%Research is needed to investigate fusion architectures that can effectively integrate information from both sensors while addressing challenges such as different data resolutions, dimensionality, and sparsity. 

\subsubsection{Handling Challenging Conditions}
%调整参数、权重；不确定性Uncertainty and Probabilistic Fusion: 
%Adaptive Fusion
%\subsubsection{Fusion in Challenging Conditions}
Fusion in challenging conditions refers to the ability of a sensor fusion system to maintain reliable and accurate perception even in adverse environments, where individual sensors may face difficulties. In such conditions, sensor measurements are noisy, degraded, or influenced by poor visibility, adverse weather and object occlusions \cite{guan2023achelous, guan2023efficient, guan2024mask, guan2023achelous++}.
Therefore, sensor fusion aims to leverage the complementary strengths of different sensors to compensate for the limitations of one sensor with the capabilities of another. Developing robust fusion algorithms that handle varying environmental conditions and adaptively combine sensor data is critical for reliable perception.

Fusion in challenging conditions requires adaptive fusion strategies that dynamically adjust fusion parameters based on environmental conditions. 
Developing data weighting techniques that assign appropriate weights to sensor measurements based on reliability can enhance fusion performance under challenging conditions \cite{nweke2019data, meng2020survey}.
For example, radar sensors may be more reliable than cameras and LiDARs under adverse weather conditions. Therefore, the fusion algorithm should adaptively emphasize radar data and de-emphasize other sensors.

Effective fusion in challenging conditions also requires uncertainty estimation associated with sensor measurements. It refers to the lack of complete confidence or knowledge about the observations obtained from different sensors in the fusion process. Probabilistic fusion techniques should be incorporated to model and propagate uncertainties properly. Bayesian neural networks are notable techniques employed for uncertainty estimation, using the prior distribution of network weights to infer the posterior distribution. This enables the computation of probabilities for specific predictions, thereby enhancing the reliability of the fusion network \cite{feng2020deep, mackay1992practical}. Such techniques should enable fusion algorithms to make informed decisions considering the uncertainty levels in fused data.

%Bayesian filtering methods such as Kalman filters, particle filters, or more advanced techniques like the Gaussian Process can be employed to achieve robust fusion in the face of uncertainty. 
%

%Fusion in challenging conditions is a crucial aspect of sensor fusion research as it aims to ensure reliable and robust perception capabilities in real-world scenarios, where environmental conditions and sensor limitations can impact the performance of individual sensors. By effectively combining the strengths of different sensors and handling the challenges posed by adverse conditions, fusion systems can enhance the overall perception performance and provide more reliable situational awareness.

%One challenge and potential research direction in fusion architectures is handling uncertainties associated with different sensors. It refers to the lack of complete confidence or knowledge about the observations obtained from different sensors in the fusion process. Uncertainty arises from various factors such as sensor noise, missing data, or inconsistent sensor measurements. Correlating the uncertainties of radar, camera, and LiDAR measurements can aid in developing more reliable fusion algorithms.
%In fusion network architecture, dealing with uncertainty involves modeling and representing the uncertainty of sensor inputs and propagating it through the network to generate reliable and robust fused outputs. 

\section{Conclusion}\label{sec:Conclusion}
In conclusion, this review explores five radar data representations (i.e., ADC signal, radar tensor, point cloud, grid map, and micro-Doppler signature) in autonomous driving. 
Through an in-depth study of radar operating principles and signal processing, we reveal the generation process, applications, as well as the advantages and limitations of these representations, providing valuable insights for the continued development of autonomous driving technology. 
By investigating advanced methods for different data representations, we aim to gain insights into the characteristics of different algorithms as well as emerging trends for radar perception.

Through analyzing various radar data representations, we thoroughly discuss the key challenges and propose potential research directions for radar perception.
In general, radar perception is progressing towards data representations containing rich information. On the one hand, representations like ADC signals and radar tensors offer increased potential information, thereby holding significant value for radar perception. On the other hand, new 4D radar sensors bring forth denser point clouds and higher resolutions, representing a noteworthy trend in autonomous driving. 
In terms of radar perception networks, the incorporation of transformer architectures, attention mechanisms, and BEV features provides valuable research directions to improve the perception performance.
Moreover, emerging radar-based sensor fusion works are also hot topics in autonomous driving and are expected in future works. 
Above all, we hope our review serves as a valuable reference for both researchers and practitioners in developing robust radar perception, making our vehicles and transportation systems safer and more efficient.

\section*{Acknowledgment}
This research was funded by Suzhou Municipal Key Laboratory for Intelligent Virtual Engineering (SZS2022004), Research Development Fund of XJTLU (RDF-19-02-23), XJTLU AI University Research Centre, Jiangsu Province Engineering Research Centre of Data Science and Cognitive Computation at XJTLU and SIP AI innovation platform (YZCXPT2022103). 
This work received financial support from Jiangsu Industrial Technology Research Institute (JITRI) and Wuxi National Hi-Tech District (WND).

\bibliographystyle{IEEEtran}
\bibliography{others,citations}%other里包含一些特殊的

\begin{IEEEbiography}[{\includegraphics[width=1in,height=1.25in,clip,keepaspectratio]{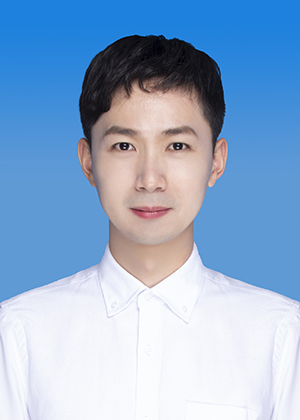}}]{Shanliang Yao} (Member, IEEE) received the B.E. degree in 2016 from the School of Computer Science and Technology, Soochow University, Suzhou, China, the M.S. degree in 2021 and Ph.D. degree in 2024 from the Faculty of Science and Engineering, University of Liverpool, Liverpool, U.K. He is currently a lecturer with the school of information engineering, Yancheng Institute Technology, Yancheng, China. His current research is centered on multi-modal perception using deep learning approach for autonomous driving. He is also interested in robotics, intelligent vehicles and intelligent transportation systems. 
\end{IEEEbiography}
%\vskip 0pt plus -1.5fil
\vskip -1cm
\begin{IEEEbiography}
[{\includegraphics[width=1in,height=1.25in,clip,keepaspectratio]{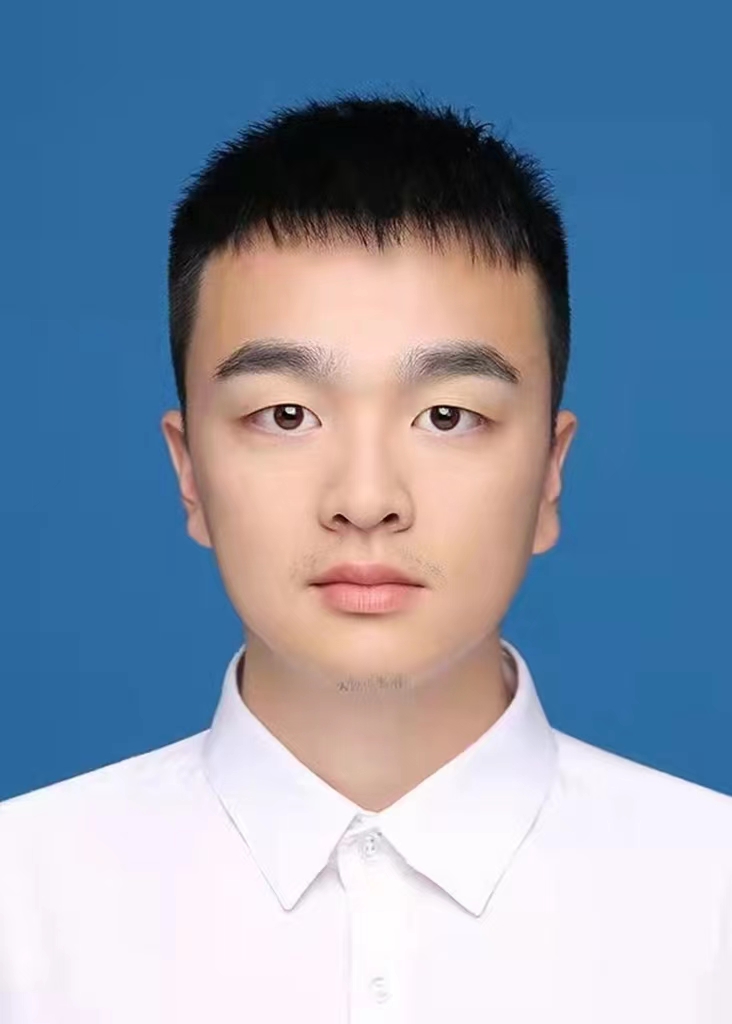}}]{Runwei Guan} (Member, IEEE) is currently a research fellow affiliated at Thrust of AI, Hong Kong University of Science and Technology (GuangZhou). He received his PhD degree from University of Liverpool in 2024 and M.S. degree in Data Science from University of Southampton in 2021. His research interests include radar perception, multi-sensor fusion, vision-language learning, lightweight neural network, multi-task learning and statistical machine learning. He serves as the peer reviewer of TITS, TNNLS, TIV, TCSVT, ITSC, ICRA, RAS, EAAI, MM, etc.
\end{IEEEbiography}
\vskip -1cm
\begin{IEEEbiography}
[{\includegraphics[width=1in,height=1.25in,clip,keepaspectratio]{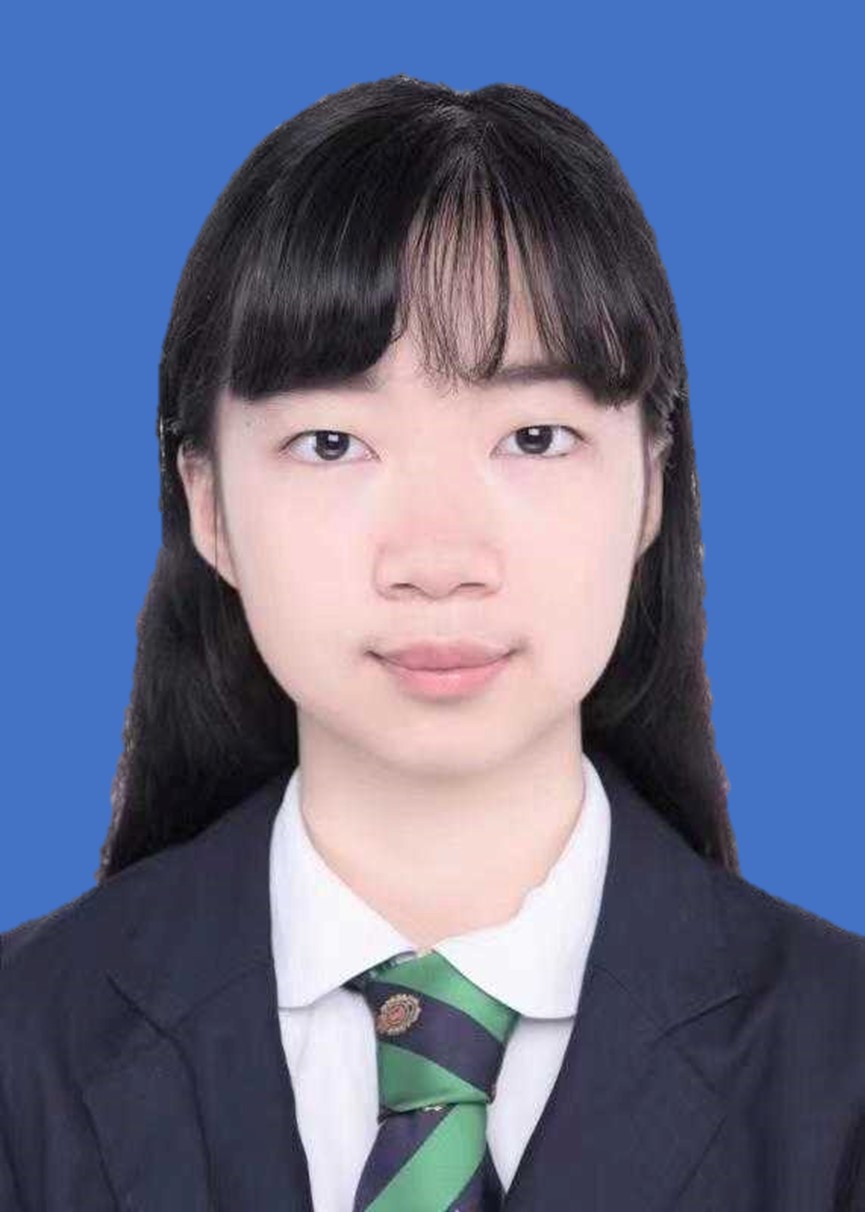}}]{Zitian Peng} received the M.S. degree in social computing in 2022 from School of Advanced Technology, Xi'an Jiaotong-Liverpool University. She is currently a Ph.D. student form University of Liverpool, Xi'an Jiaotong-Liverpool University. Her research interests include virtual reality, augmented reality, human-vehicle interaction, and reinforcement learning for gaze estimation and tracking. She is also interested in the application of digital twin in unmanned vehicles research including high-accuracy marine environment modeling, interactive DT via immersive realities, path planning algorithmic advancement and cost implication.
\end{IEEEbiography}
\vskip -1cm
\begin{IEEEbiography}
[{\includegraphics[width=1in,height=1.25in,clip,keepaspectratio]{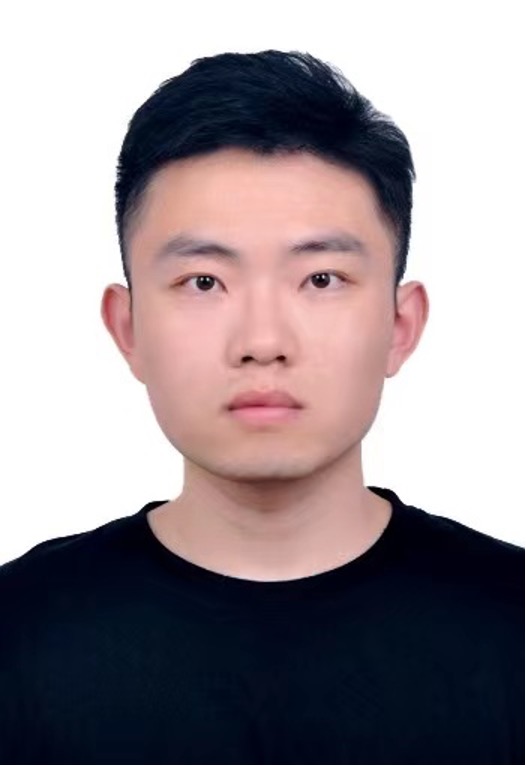}}]{Chenhang Xu} received the B.E. degree in 2018 from the Department of Electrical and Computer Engineering, Iowa State University, Ames, IA, USA, and the M.S. degree in 2020 from the Department of Electrical and Computer Engineering, Iowa State University, Ames, IA, USA. He is currently a Ph.D. student of University of Liverpool and Xi'an Jiaotong-Liverpool University. His current research is centered on multi-agent reinforcement learning cooperation. He is also interested in robotics and intelligent vehicles. 
\end{IEEEbiography}
\vskip -1cm
\begin{IEEEbiography}
[{\includegraphics[width=1in,height=1.25in,clip,keepaspectratio]{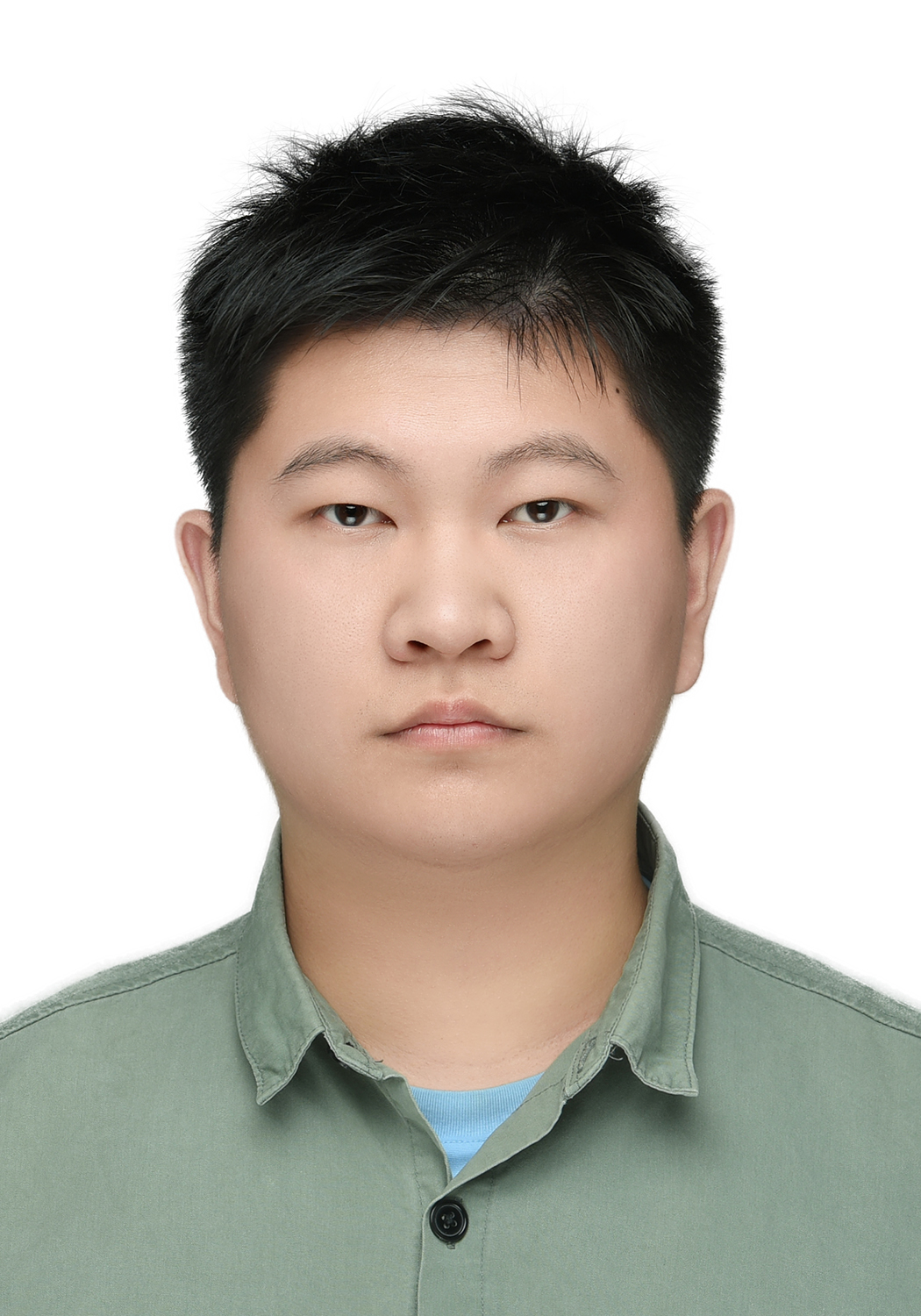}}]{Yilu Shi}  is currently an undergraduate student at Xi'an Jiaotong Liverpool University, majoring in information and computing science. He will complete his undergraduate study in 2024. His research interests include multi-modal detection, tracking and computer vision for autonomous driving. He won the third prize in the 9th ``Internet Plus" College Student Innovation and Entrepreneurship Competition in Jiangsu Province and participated the Programme and Poster Competition as a Summer Undergraduate Research Fellow in 2023.
\end{IEEEbiography}
\vspace{-1cm}
\begin{IEEEbiography}[{\includegraphics[width=1in,height=1.25in,clip,keepaspectratio]{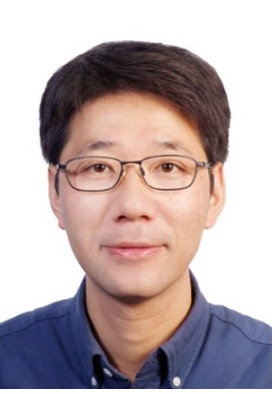}}]{Weiping Ding} (M'16-SM'19) received the Ph.D. degree in Computer Science, Nanjing University of Aeronautics and Astronautics, Nanjing, China, in 2013. In 2016, He was a Visiting Scholar at National University of Singapore, Singapore. From 2017 to 2018, he was a Visiting Professor at University of Technology Sydney, Australia. He is a Full Professor with the School of Information Science and Technology, Nantong University, Nantong, China, and also the supervisor of Ph.D postgraduate by the Faculty of Data Science at City University of Macau, China. His main research directions involve deep neural networks, multimodal machine learning, and medical images analysis. He ranked within the top 2\% Ranking of Scientists in the World by Stanford University (2020-2023). He has published over 250 articles, including over 100 IEEE Transactions papers. His fifteen authored/co-authored papers have been selected as ESI Highly Cited Papers. He serves as an Associate Editor/Editorial Board member of IEEE Transactions on Neural Networks and Learning Systems, IEEE Transactions on Fuzzy Systems, IEEE/CAA Journal of Automatica Sinica, IEEE Transactions on Intelligent Transportation Systems, IEEE Transactions on Intelligent Vehicles, IEEE Transactions on Emerging Topics in Computational Intelligence, IEEE Transactions on Artificial Intelligence, Information Fusion, Information Sciences, Neurocomputing, Applied Soft Computing. He is the Leading Guest Editor of Special Issues in several prestigious journals, including IEEE Transactions on Evolutionary Computation, IEEE Transactions on Fuzzy Systems, and Information Fusion.
\end{IEEEbiography}
\vspace{-1cm}
\begin{IEEEbiography}
[{\includegraphics[width=1in,height=1.25in,clip,keepaspectratio]{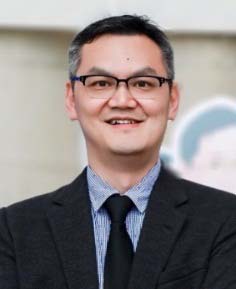}}]{Eng Gee Lim}
(Senior Member, IEEE) received the B.Eng. (Hons.) and Ph.D. degrees in Electrical and Electronic Engineering (EEE) from Northumbria University, Newcastle, U.K., in 1998 and 2002,
respectively. He worked for Andrew Ltd., Coventry, U.K., a leading communications systems company from 2002 to 2007. Since 2007, he has been with Xi'an Jiaotong-Liverpool University, Suzhou, China, where he was the Head of the EEE Department, and the University Dean of research and graduate studies. He is currently the School Dean of Advanced Technology, the Director of the AI University Research Centre, and a Professor with the Department of EEE. He has authored or coauthored over 100 refereed international journals and conference papers. His research interests are artificial intelligence (AI), robotics, AI+ health care, international standard (ISO/IEC) in robotics, antennas, RF/microwave engineering, EM measurements/simulations, energy harvesting, power/energy transfer, smart-grid communication, and wireless communication networks for smart and green cities. He is a Charted Engineer and a fellow of The Institution of Engineering and Technology (IET) and Engineers Australia. He is also a Senior Fellow of Higher Education Academy (HEA).
\end{IEEEbiography}
\vspace{-1cm}
\begin{IEEEbiography}
[{\includegraphics[width=1in,height=1.25in,clip,keepaspectratio]{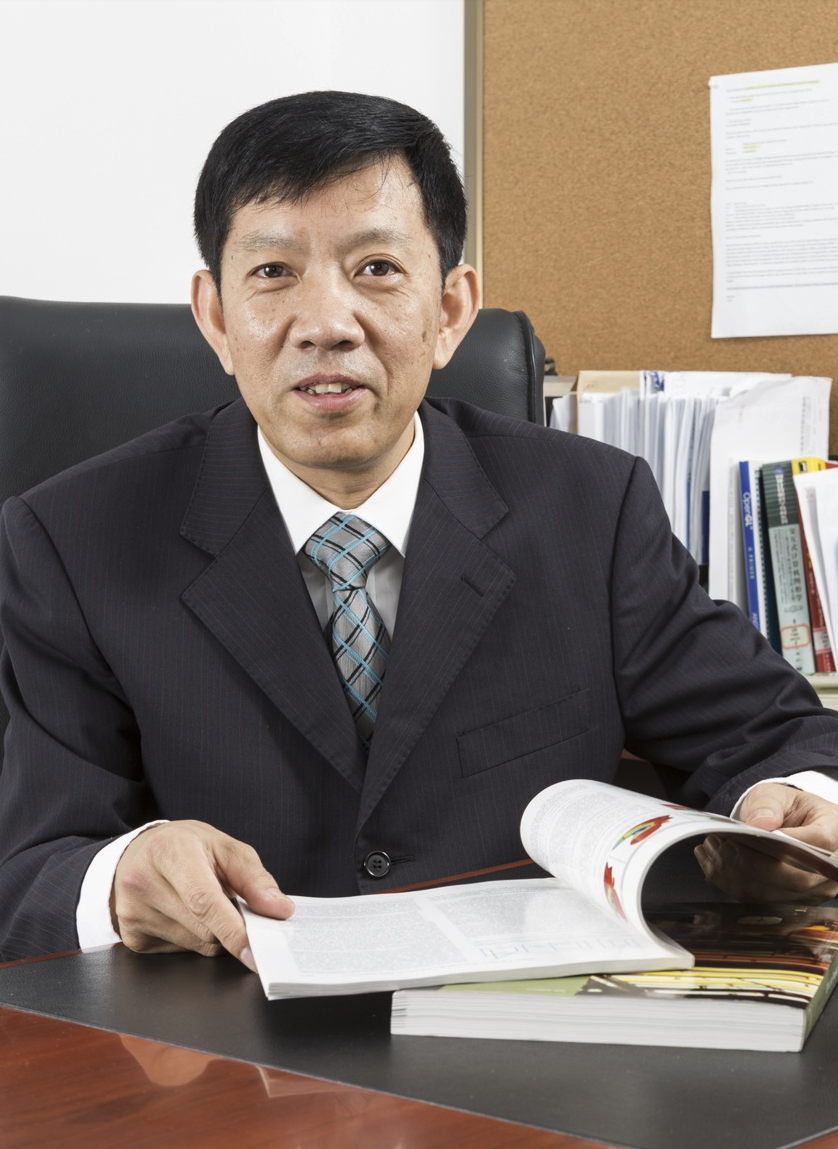}}]{Yong Yue}
Fellow of Institution of Engineering and Technology (FIET), received the B.Eng. degree in mechanical engineering from Northeastern University, Shenyang, China, in 1982, and the Ph.D. degree in computer aided design from Heriot-Watt University, Edinburgh, U.K., in 1994. He worked in the industry for eight years and followed experience in academia with the University of Nottingham, Cardiff University, and the University of Bedfordshire, U.K. He is currently a Professor and Director with the Virtual Engineering Centre, Xi'an Jiaotong-Liverpool University, Suzhou, China. His current research interests include computer graphics, virtual reality, and robot navigation.
\end{IEEEbiography}
\vspace{-1cm}
\begin{IEEEbiography}[{\includegraphics[width=1in,height=1.25in,clip,keepaspectratio]{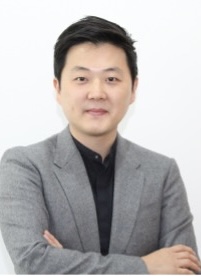}}]{Hyungjoon Seo} (Member, IEEE) received the bachelor's degree in civil engineering from Korea University, Seoul, South Korea, in 2007, and the Ph.D. degree in geotechnical engineering from Korea University in 2013. In 2013, he worked as a research professor in Korea University. He served as a visiting scholar at University of Cambridge, Cambridge, UK, and he worked for engineering department in University of Cambridge as a research associate from 2014 to 2016. In August 2016, he got an assistant professor position in the Department of Civil Engineering at the Xi'an Jiaotong Liverpool University (XJTLU), China. He has been an assistant professor at the University of Liverpool, UK, from 2020. His research interests are monitoring using artificial intelligence and SMART monitoring system for infrastructure, soil-structure interaction (tunneling, slope stability, pile), Antarctic survey and freezing ground. 
Hyungjoon is the director of the CSMI (Centre for SMART Monitoring Infrastructure), CSMI is collaborating with University of Cambridge, University of Oxford, University of Bath, UC Berkeley University, Nanjing University, and Tongji University on SMART monitoring. He presented a keynote speech at the 15th European Conference on Soil Mechanics and Geotechnical Engineering in 2015. He is currently appointed editor of the CivilEng journal and organized two international conferences. He has published more than 50 scientific papers including a book on Geotechnical Engineering and SMART monitoring. 
\end{IEEEbiography}
\vspace{-1cm}
\begin{IEEEbiography}[{\includegraphics[width=1in,height=1.25in,clip,keepaspectratio]{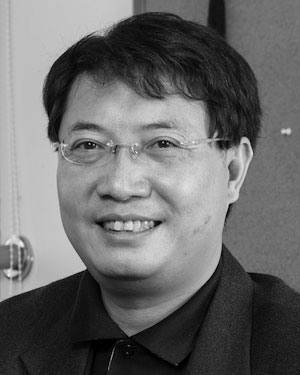}}]{Ka Lok Man}
(Member, IEEE) received the Dr. Eng. degree in electronic engineering from the Politecnico di Torino, Turin, Italy, in 1998, and the Ph.D. degree in computer science from Technische Universiteit Eindhoven, Eindhoven, The Netherlands, in 2006. He is currently a Professor in Computer Science and Software Engineering with Xi'an Jiaotong-Liverpool University, Suzhou, China. His research interests include formal methods and process algebras, embedded system design and testing, and photovoltaics.
\end{IEEEbiography}
\vspace{-1cm}
\begin{IEEEbiography}[{\includegraphics[width=1in,height=1.25in,clip,keepaspectratio]{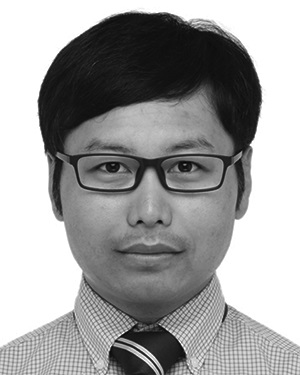}}]{Jieming Ma} received the M.Sc. degree in advanced microelectronic systems engineering from the University of Bristol, UK, in 2010, and received the Ph.D. degree in computer science from the University of Liverpool, UK, in 2014. He is currently working as an Associate Professor at the Xi'an Jiaotong-Liverpool University, China. His research interests include intelligent optimization, machine learning and applications in renewable energy systems.
\end{IEEEbiography}
\vspace{-1cm}
\begin{IEEEbiography}
[{\includegraphics[width=1in,height=1.25in,clip,keepaspectratio]{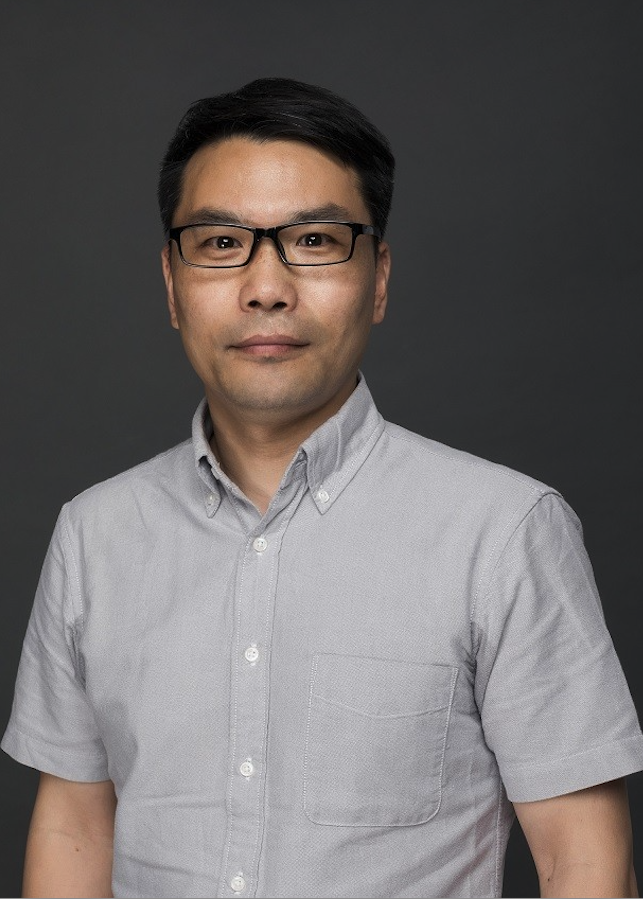}}]{Xiaohui Zhu}
(Member, IEEE) received his Ph.D. from the University of Liverpool, UK in 2019. He is currently an associate  professor, Ph.D. supervisor and Programme Director with the Department of Computing, School of Advanced Technology, Xi'an Jiaotong-Liverpool University. He focuses on advanced techniques related to autonomous driving, including sensor-fusion perception, fast path planning, autonomous navigation and multi-vehicle collaborative scheduling. 
\end{IEEEbiography}
\vspace{-1cm}
\begin{IEEEbiography}[{\includegraphics[width=1in,height=1.25in,clip,keepaspectratio]{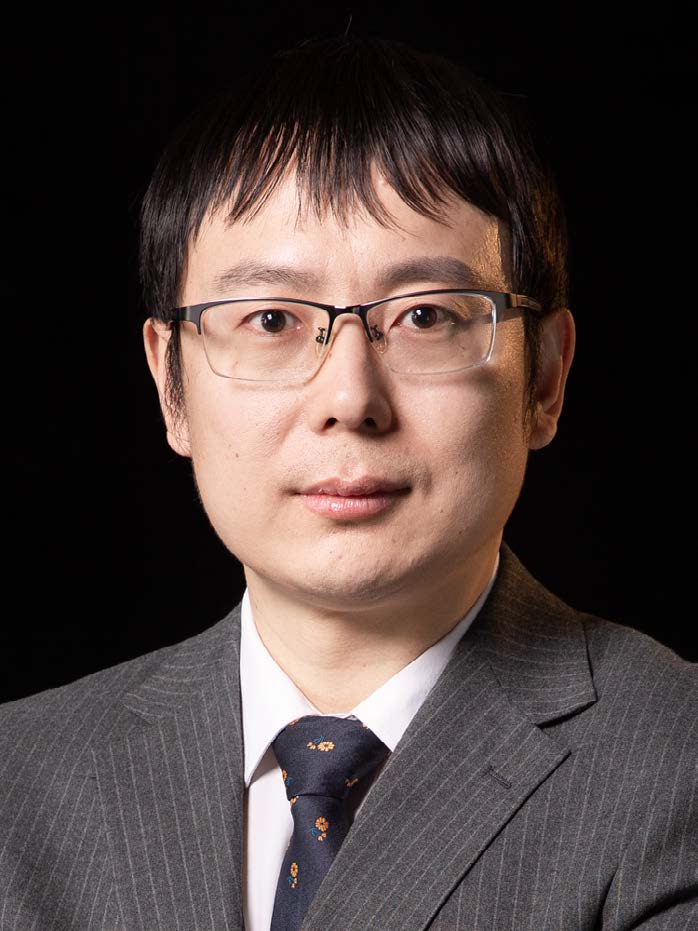}}]{Yutao Yue} (Member, IEEE) is an associate professor at the Artificial Intelligence Thrust and Intelligent Transportation Thrust of Hong Kong University of Science and Technology (Guangzhou). He received his Bachelor’s degree from the University of Science and Technology of China, and Master and PhD degree from Purdue University. He has a dual background in academia and industry, as the team leader of Guangdong Province Introduced Innovation Scientific Research Team, senior scientist of Kuang-Chi Group, and the founder of the Institute of Deep Perception Technology of JITRI. His research interests include multimodal perception fusion, machine consciousness, artificial general intelligence, causal emergence, etc. He has been engaged in scientific research and technology industrialization for over 20 years. He has co-invented 354 granted Chinese patents, 18 USA patents, and 7 EU patents. He has led 6 major research projects with a total funding of nearly 130 million RMB. He has published over 60 papers, advised 13 postdoc research fellows, and received multiple awards including Wu Wenjun Artificial Intelligence Science and Technology Award.
\end{IEEEbiography}

\end{document}